%% file: main.tex
\documentclass[11pt]{article}

\usepackage{fullpage}
\usepackage[ruled, linesnumbered, vlined, commentsnumbered]{algorithm2e}
\usepackage{graphicx,subfig} 
\usepackage{amsfonts,amssymb,amsmath,amsthm,amsopn}	
\usepackage{hhline,booktabs,colortbl,multirow,tabularx,threeparttable} 

\usepackage{enumerate}
\usepackage{authblk}
\usepackage{footnote}
\usepackage{hyperref}
\usepackage{prettyref}
\usepackage{cite}
\usepackage{setspace}
\usepackage{color}
\usepackage{xcolor}  
\usepackage{scrpage2}
\usepackage{geometry}

\usepackage{tikz}
\usepackage{pgfplots}
\usetikzlibrary{positioning,shapes,shadows,arrows,calc}
\tikzstyle{component}=[rectangle, draw=black, rounded corners, fill=blue!40, drop shadow, text centered, anchor=north, text=white, minimum height=1cm]
\tikzstyle{arrow}=[->, thick]

\pgfplotsset{compat=1.12}
\usetikzlibrary{intersections}
\usetikzlibrary{pgfplots.statistics}
\usepgfplotslibrary{fillbetween}

\definecolor{myblue}{RGB}{34,31,217}
\definecolor{mycyan}{gray}{.7}
\definecolor{Gray}{gray}{0.9}

\newtheorem{definition}{Definition}

\newcommand{\pref}{\prettyref}

\newrefformat{fig}{Fig.~\ref{#1}}
\newrefformat{tab}{Table~\ref{#1}}
\newrefformat{sec}{Section~\ref{#1}}
\newrefformat{app}{Appendix~\ref{#1}}
\newrefformat{alg}{Algorithm~\ref{#1}}
\newrefformat{property}{Property~\ref{#1}}
\newrefformat{theorem}{Theorem~\ref{#1}}
\newrefformat{corollary}{Corollary~\ref{#1}}
\newrefformat{proposition}{Proposition~\ref{#1}}
\newrefformat{def}{Definition~\ref{#1}}
\newrefformat{eq}{equation~(\ref{#1})}

\usepackage{lscape}

\begin{document}

\title{\vspace{-1ex}\LARGE\textbf{Decomposition Multi-Objective Evolutionary Optimization: From State-of-the-Art to Future Opportunities}\footnote{This manuscript is submitted for potential publication. Reviewers can use this version in peer review.}}

\author{\normalsize Ke Li}
\affil{\normalsize Department of Computer Science\\ University of Exeter, North Park Road, Exeter, EX4 4QF, UK}
\affil[$\ast$]{\normalsize Email: \texttt{k.li@exeter.ac.uk}}

\date{}
\maketitle

\vspace{-3ex}
{\normalsize\textbf{Abstract: } }Decomposition has been the mainstream approach in the classic mathematical programming for multi-objective optimization and multi-criterion decision-making. However, it was not properly studied in the context of evolutionary multi-objective optimization until the development of multi-objective evolutionary algorithm based on decomposition (MOEA/D). In this article, we present a comprehensive survey of the development of MOEA/D from its origin to the current state-of-the-art approaches. In order to be self-contained, we start with a step-by-step tutorial that aims to help a novice quickly get onto the working mechanism of MOEA/D. Then, selected major developments of MOEA/D are reviewed according to its core design components including weight vector settings, subproblem formulations, selection mechanisms and reproduction operators. Besides, we also overviews some further developments for constraint handling, computationally expensive objective functions, preference incorporation, and real-world applications. In the final part, we shed some lights on emerging directions for future developments.


\input{introduction}

\input{background}

\input{review}

\input{advanced}

\input{applications}

\input{emerging}

\input{conclusion}

\section*{Acknowledgment}
K. Li was supported by UKRI Future Leaders Fellowship (MR/S017062/1), EPSRC (2404317) and Amazon Research Awards.

\bibliographystyle{IEEEtran}
\bibliography{IEEEabrv,moead}

\end{document}

%% file: introduction.tex

\section{Introduction}
\label{sec:introduction}

Real-world problems, such as machine learning~\cite{SenerK18}, data mining~\cite{RibeiroZMHLV14}, material science~\cite{AttiaGJSMLCCPYH20}, cheminformatics~\cite{LiHWZZL20}, aerospace~\cite{HornbyLL11}, software engineering~\cite{ChenLBY18}, finance and economics~\cite{PonsichJC13}, usually require solutions to simultaneously meet multiple objectives, known as multi-objective optimization problems (MOPs). These objectives conflict with each other where an improvement in one objective can lead to a detriment of other(s). There does not exist a global optimum that optimizes all objectives but a set of trade-off solutions. Due to the population-based property, evolutionary computation (EC) has been widely recognized as a major approach for multi-objective optimization (MO). Over the past three decades and beyond, many efforts have been dedicated to developing evolutionary multi-objective optimization (EMO) algorithms, which can be classified as dominance-, indicator-, and decomposition-based frameworks. In particular, fast non-dominated sorting genetic algorithm (NSGA-II)~\cite{DebAPM02}, indicator-based evolutionary algorithm~\cite{ZitzlerK04} and multi-objective evolutionary algorithm based on decomposition (MOEA/D)~\cite{ZhangL07} are representative algorithms respectively.

\begin{figure}[t!]
	\centering
    \includegraphics[width=.6\linewidth]{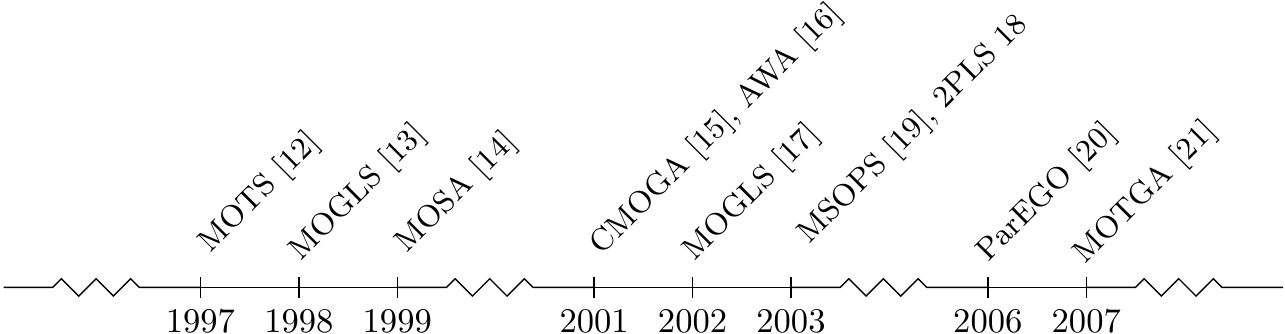}
    \caption{Development trajectories of MOEA/D from the past to the state-of-the-art.}
    \label{fig:timeline}
\end{figure}

Different from the other two frameworks, which are rooted in the EC community, decomposition has been the mainstream approach in the classic mathematical programming for MO and multi-criterion decision-making (MCDM)~\cite{Miettinen99}. Its basic idea is to aggregate different objectives into a scalarizing function by using a dedicated weight vector. An MOP is thus transformed into a single-objective optimization subproblem, the optimum of which is a Pareto-optimal solution of the original MOP. However, traditional mathematical programming approaches are notorious for heavily relying on the derivative information which is unlikely to access in real-world black-box scenarios. Moreover, only one Pareto-optimal solution can be obtained at a time by using mathematical programming in MO thus compromising the effectiveness and flexibility for exploring various trade-off solutions at a time. Before MOEA/D, there have been some attempts to incorporate decomposition techniques into EMO, such as~\cite{Hansen97,IshibuchiM98,UlunguTFT99,MurataIG01,JinOS01,Jaszkiewicz02b,PaqueteS03,Hughes05,Knowles06,AlvesA07} listed in a chronological timeline shown in~\pref{fig:timeline}. However, since the weight vectors therein are generated in a random manner, these prior algorithms do not have a principled way to maintain population diversity. Meanwhile, the randomly generated weight vectors make the search directions keep on alternating thus leading to a less effective search process~\cite{GiagkiozisPF13}. Furthermore, none of those prior algorithms take the collaboration among neighboring subproblems into consideration thus they are hardly capable of tackling problems with complex properties~\cite{LiZ09}. After MOEA/D, decomposition-based framework has become the most popular one in the EMO community.

This paper aims to provide a comprehensive survey of the up-to-date studies on MOEA/D. We are ambitious to be self-contained so that a novice can quickly get onto the working mechanism of MOEA/D through a step-by-step tutorial as in~\pref{sec:background}. Selected major developments of MOEA/D are surveyed in~\pref{sec:main_review} according to its core design components, i.e., weight vector settings, subproblem formulations, selection mechanisms and reproduction operators. Besides, \pref{sec:advanced} overviews some further developments for constraint handling, computationally expensive objective functions, preference incorporation and real-world applications. Some emerging directions for future developments of MOEA/D are discussed in~\pref{sec:others} and~\pref{sec:conclusions} concludes this paper at the end.

%% file: background.tex

\section{Background Knowledge}
\label{sec:background}

In this section, we start with some basic concepts closely relevant to this paper. Thereafter, we give a gentle tutorial of the working mechanisms of a vanilla MOEA/D.

\subsection{Basic Concepts}
\label{sec:concepts}

The MOP considered in this paper is mathematically defined as:
\begin{equation}
\begin{array}{l l}
\min \quad\quad\quad\, \mathbf{F}(\mathbf{x})=(f_{1}(\mathbf{x}),\cdots,f_{m}(\mathbf{x}))^{T}\\
\mathrm{subject\ to} \;\;\, g_j(\mathbf{x})\geq a_j,\quad j=1,\cdots,q\\
\mathrm{\ } \quad\quad\quad\quad\;\;\, h_k(\mathbf{x})=b_k,\;\;\, k=q+1,\cdots,\ell\\
\mathrm{\ } \quad\quad\quad\quad\;\;\, \mathbf{x} \in\Omega
\end{array},
\label{MOP}
\end{equation}
where $\mathbf{x}=(x_1,\ldots,x_n)^T$ is a solution, and $\Omega=[x_i^L,x_i^U]^n\subseteq\mathbb{R}^n$ defines the search (or decision variable) space. $\mathbf{F}:\Omega\rightarrow\mathbb{R}^m$ constitutes $m$ conflicting objective functions, and $\mathbb{R}^m$ is the objective space. $g_j(\mathbf{x})$ and $h_k(\mathbf{x})$ are the $j$-th inequality and $k$-th equality constraints respectively.

\begin{definition}
Given two feasible solutions $\mathbf{x}^1$ and $\mathbf{x}^2$, $\mathbf{x}^1$ is said to \textit{Pareto dominate} $\mathbf{x}^2$ (denoted as $\mathbf{x}^1\preceq\mathbf{x}^2$) if and only if $f_i(\mathbf{x}^1)\leq f_i(\mathbf{x}^2)$, $\forall i\in\{1,\cdots,m\}$ and $\exists j\in\{1,\cdots,m\}$ such that $f_i(\mathbf{x}^1)<f_i(\mathbf{x}^2)$.
\end{definition}

\begin{definition}
A solution $\mathbf{x}^\ast\in\Omega$ is \textit{Pareto-optimal} with respect to (\ref{MOP}) if $\nexists\mathbf{x}\in\Omega$ such that $\mathbf{x}\preceq\mathbf{x}^{\ast}$.
\end{definition}

\begin{definition}
The set of all Pareto-optimal solutions is called the \textit{Pareto-optimal set} (PS). Accordingly, $PF=\{\mathbf{F}(\mathbf{x})|\mathbf{x}\in PS\}$ is called the \textit{Pareto-optimal front} (PF).
\end{definition}

\begin{definition}
The \textit{ideal objective vector} $\mathbf{z}^\ast=(z^\ast_1,\cdots,z^\ast_m)^T$ is constructed by the minimum of each objective functions, i.e., $z^\ast_i=\min_{\mathbf{x}\in\Omega}f_i(\mathbf{x})$, $i\in\{1,\cdots,m\}$.
\end{definition}

\begin{definition}
The \textit{nadir objective vector} $\mathbf{z}^{nad}=(z^{nad}_1,\cdots,z^{nad}_m)^T$ is constructed by the worst objective functions of the PF, i.e., $z^{nad}_i=\max_{\mathbf{x}\in PS}f_i(\mathbf{x})$, $i\in\{1,\cdots,m\}$.
\end{definition}

\begin{figure}[t]
	\centering
    \includegraphics[width=\linewidth]{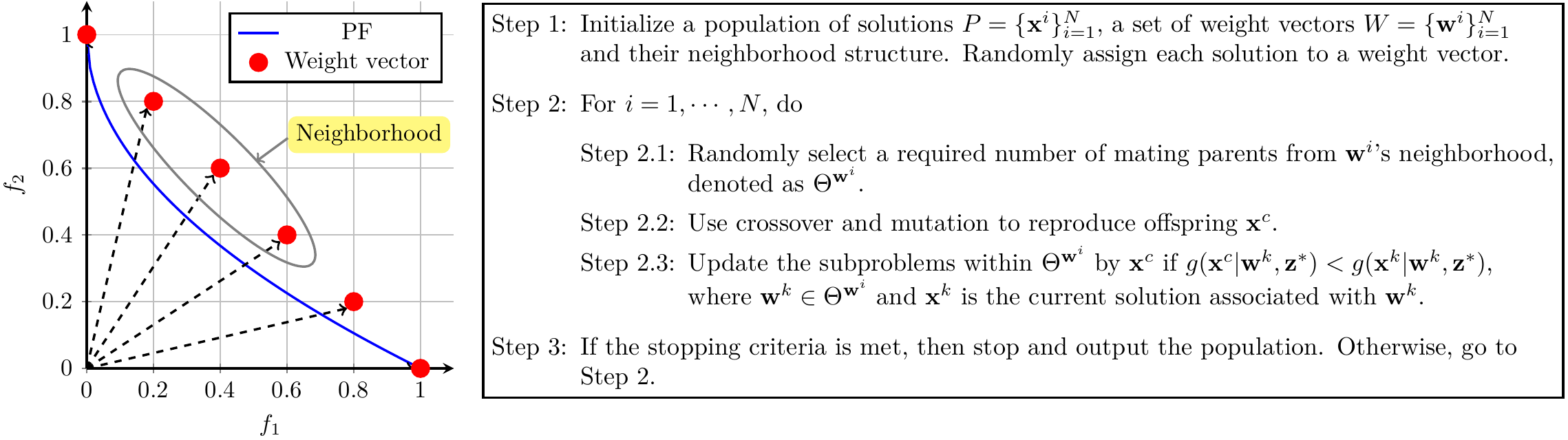}
    \caption{A schematic illustration and a three-step procedure of vanilla MOEA/D.}
    \label{fig:vanilla_moead}
\end{figure}
\vspace{-1em}

\subsection{Working Mechanisms of Vanilla MOEA/D}
\label{sec:moead}

\pref{fig:vanilla_moead} gives a schematic illustration and a three-step procedure of vanilla MOEA/D. It consists of two major functional components, i.e., \textit{decomposition} and \textit{collaboration}, the mechanisms of which are delineated as follows.

\subsubsection{Decomposition}
\label{sec:decomposition}


Under some mild continuous conditions, a Pareto-optimal solution of a MOP is an optimal solution of a scalar optimization problem whose objective function is an aggregation of all individual objectives~\cite{Miettinen99}. The principle of \textit{decomposition} is to transform the task of approximating the PF into a number of scalarized subproblems. There have been various approaches for constructing such subproblems~\cite{Miettinen99}, which will be further discussed in~\pref{sec:subproblems}. Here we introduce three most widely used ones in the decomposition-based EMO community. 
\begin{itemize}
\item\underline{\textit{Weighted Sum (WS)}}: it is a convex combination of all individual objectives. Let $\mathbf{w}=(w_1,\cdots,w_m)^T$ be a weight vector where $w_i\geq 0$, $\forall i\in\{1,\cdots,m\}$ and $\sum_{i=1}^{m}w_i=1$, a WS is formulated as:
\begin{equation}
\min_{\mathbf{x}\in\Omega}\quad g^{\mathtt{ws}}(\mathbf{x}|\mathbf{w})=\sum_{i=1}^{m}w_{i}f_i(\mathbf{x}).
\label{equation:WS}
\end{equation}

\item\underline{\textit{Weighted Tchebycheff (TCH)}}: in this approach, the scalar optimization problem is formulated as:
\begin{equation}
\min_{\mathbf{x}\in\Omega}\quad g^{\mathtt{tch}}(\mathbf{x}|\mathbf{w},\mathbf{z}^{\ast})=\max\limits_{1\leq i\leq m}\big\{w_i|f_i(\mathbf{x})-z_{i}^{\ast}|\big\},
\label{equation:TCH}
\end{equation}
where $\mathbf{z}^\ast$ is the approximated ideal objective vector estimated by using the current population.

\item\underline{\textit{Penalty-baed Boundary Intersection (PBI)}}: this is a variant of the normal-boundary intersection method~\cite{DasD98}, whose equality constraint is handled by a penalty function. Formally, it is defined as:
\begin{equation}
\min_{\mathbf{x}\in\Omega}\quad g^{\mathtt{pbi}}(\mathbf{x}|\mathbf{w},\mathbf{z}^{\ast})=d_1+\theta d_2,
\label{eq:pbi}
\end{equation}
where $\theta>0$ is a user-defined penalty parameter, $d_1=\|(\mathbf{F}(\mathbf{x})-\mathbf{z}^{\ast})^T\mathbf{w}\|/\|\mathbf{w}\|$ and $d_2=\|\mathbf{F}(\mathbf{x})-(\mathbf{z}^{\ast}+d_1\mathbf{w})\|$. 
\end{itemize}

\begin{figure}[t!]
	\centering
    \includegraphics[width=\linewidth]{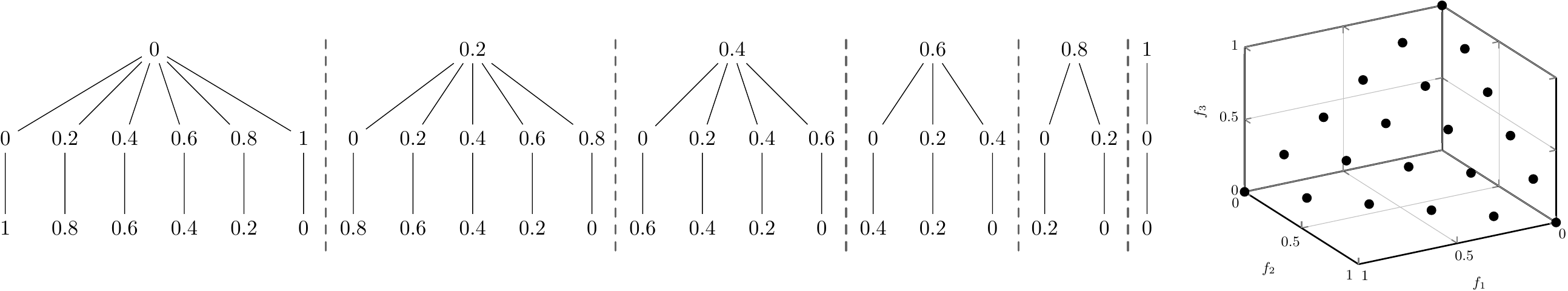}
    \caption{An illustrative example of weight vector generation in the 3-dimensional space by using the Das and Dennis's method~\cite{DasD98}.}
    \label{fig:NBI_weights}
\end{figure}

Note that MOEA/D applies the Das and Dennis's method~\cite{DasD98} to specify a set of weight vectors evenly distributed along a unit simplex. Its basic idea is to divide each axis into $H>0$ equally spaced segments. The weight vectors are constituted by picking up a sliced coordinate along each axis in an iterative manner. This Das and Dennis's method can generate ${H+m-1}\choose{m-1}$ weight vectors in total. Each weight vector corresponds to a unique subproblem in MOEA/D. \pref{fig:NBI_weights} gives an illustrative example of $21$ weight vectors generated in a three-dimensional space where $H=5$.

\subsubsection{Collaboration}
\label{sec:collaboration}

Different from the classic mathematical programming method for multi-objective optimization, in which only one subproblem is considered at a time, MOEA/D applies a population-based meta-heuristic to concurrently solve all subproblems in a collaborative manner. Each subproblem is associated with its best-so-far solution during the evolutionary problem-solving process. The basic assumption behind this collaboration mechanism is that neighboring subproblems are more likely to share similar properties, e.g., similar objective functions and/or optimal solutions. In particular, the neighborhood of a subproblem is determined by the Euclidean distance of its corresponding weight vector with respect to the others. By this means, solutions associated with the suproblems lying in the same neighborhood can collaborate with each other to push the evolution forward by sharing and exchanging elite information. More specifically, the mating parents are picked up from the underlying neighborhood for offspring reproduction while the newly generated offspring is used to update the subproblem(s) within the same neighborhood in mating selection. Note that the solution associated with a subproblem can be updated/replaced by the new one if it has a better fitness value.


%% file: review.tex

\section{Selected Developments on Major Components of MOEA/D}
\label{sec:main_review}

\input{weights}

\input{subproblems}

\input{selection}

\input{reproduction}

%% file: weights.tex

In this section, we plan to overview some selected developments in the major components of MOEA/D including the weight vector settings, subproblem formulations, selection and reproduction.

\subsection{Weight Vector Setting}
\label{sec:weight}

\begin{table}[htbp]
\centering
\caption{Selected works for weight vector settings in MOEA/D.}
\label{tab:weight_vector}
\resizebox{\textwidth}{!}{ 
    \begin{tabular}{c|c|c}
    \hline
        \textsc{Subcategory} & \textsc{Algorithm Name} & \textsc{Core Technique} \\
        \hline\hline
        \multirow{1}[8]{*}{Fixed methods} & MOGLS~\cite{IshibuchiM98,Jaszkiewicz02a,Jaszkiewicz02b}, MOEA/D-RW~\cite{LiDDZ15} & Random method \\
        \cline{2-3}          & NSGA-III~\cite{DebJ14}, MOEA/DD~\cite{LiDZK15} & Multi-layer method \\
        \cline{2-3}          & UMOEA/D~\cite{TanJLW12,TanJ13}, UMODE/D~\cite{TanJLW13}  & Uniform design \\
        \cline{2-3}          & Riesz $s$-energy function~\cite{BlankDDBS21}, RMA~\cite{JinOS01} & Others \\
        \hline\hline
        \multirow{1}[6]{*}{Model-based adaptation methods} & T-MOEA/D~\cite{LiuGC10}, $pa\lambda$-MOEA/D~\cite{JiangCZO11}, $apa$-MOEA/D~\cite{JiangZO11}, DMOEA/D~\cite{GuLT12} & Polynomial model \\
        \cline{2-3}          & MOEA/D-LTD~\cite{WuKJLZ17,WuLKZZ19} & Parametric model \\
        \cline{2-3}          & MOEA/D-SOM~\cite{GuC18}, MOEA/D-GNG~\cite{LiuIMN20,LiuJHRY20} & Neural networks \\
        \hline\hline
        \multirow{3}[16]{*}{Delete-and-add} & MOEA/D-AWA~\cite{QiMLJSW14}, MOEA/D-URAW~\cite{FariasBBA18,FariasA19}, AdaW~\cite{LiY20} & \multirow{1}[4]{*}{Less crowded or promising areas} \\
        \cline{2-2}          & CLIA~\cite{GeZSWTZC19}, MOEA/D-AM2M~\cite{LiuCZD16,LiuCZD18}  &  \\
        \cline{2-3}          & RVEA*~\cite{ChengJOS16}, \cite{ZhaoZZDY18}, iRVEA~\cite{GuCLL18}, OD-RVEA~\cite{LiuJHR19}, EARPEA~\cite{ZhouDZLM18}, g-DBEA~\cite{AsafuddoulaSR18} & Away from the promising areas \\
        \cline{2-3}          & A-NSGA-III~\cite{JainD14}, A$^2$-NSGA-III~\cite{JainD13}, AMOEA/D~\cite{CamachoPBI19}, MOEA/D-TPN~\cite{ZhangZLCC18},  & \multirow{2}[4]{*}{Regulated neighborhood} \\
        \cline{2-2}          & TPEA-PBA~\cite{ZhuCFS16}, MaOEA/D-2ADV~\cite{CaiMF18}, EMOSA~\cite{LiS08,LiL11}, AWA~\cite{XuZZZ19} &  \\
        \cline{2-2}          & AWA-SSCWA~\cite{HamadaNKO11}, MOEA/D-HD~\cite{ShiodaO15} & \multirow{2}[6]{*}{Local niches of non-dominated solutions} \\
        \cline{2-3}          & AWD-MOEA/D~\cite{GuoWW15}, FV-MOEA/D~\cite{JiangFHNOZT16}, AR-MOEA~\cite{TianCZCJ18}, MOEA-ABD~\cite{ZhangTLG18} &  \\
        \cline{2-2}          & E-IM-MOEA~\cite{LinLJ18}, MOEA/D-AWVAM~\cite{HaradaHH17} &  \\
        \hline
        \end{tabular}%
}
\label{tab:addlabel}%
\end{table}%

\begin{figure}[t!]
	\centering
    \includegraphics[width=\linewidth]{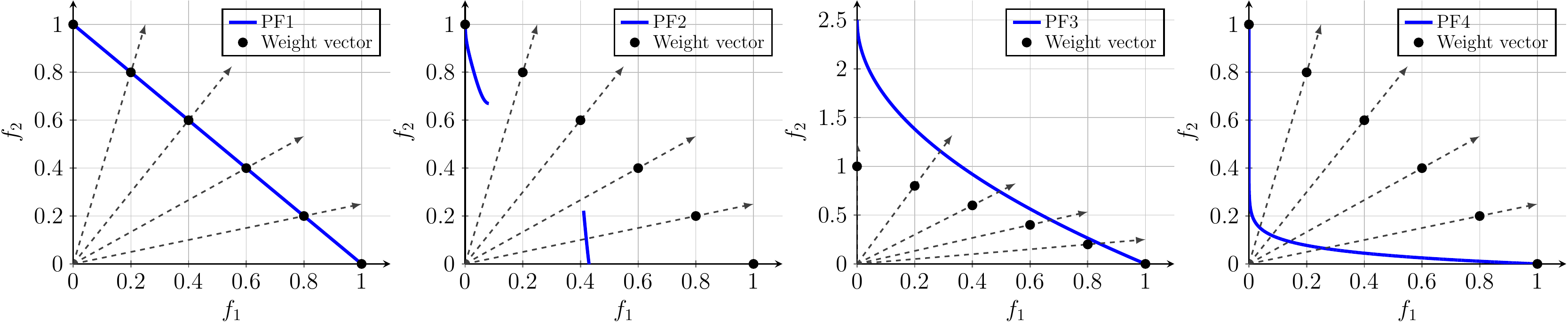}
    \caption{MOEA/D obtains evenly distributed solutions by using evenly distributed weight vectors merely when the PF is a simplex. Otherwise, some weight vectors do not have Pareto-optimal solutions like PF2; or the distribution of the corresponding Pareto-optimal solutions is highly biased like PF3 and PF4.}
    \label{fig:weight_discussion}
\end{figure}

A weight vector plays as the carrier of a subproblem in MOEA/D while the setting of weight vectors is essential to achieve a well balance between convergence and diversity~\cite{IshibuchiSMN17}. However, it is arguable that using a set of evenly distributed weight vectors can be detrimental if the PF is not a regular simplex, such as disconnected, badly scaled and strong convex shapes shown in~\pref{fig:weight_discussion}. To deal with this issue, there have been a significant amount of works on proactively adapting the setting of weight vectors. This subsection overviews some selected developments along this line of research according to the type of heuristics for adaptation, as summarized in~\pref{tab:weight_vector}.

\vspace{-1em}

\subsubsection{Fixed methods}
\label{sec:fixed_method}

Instead of progressively tuning the weight vectors, this type of method uses a fixed strategy to set the weight vectors according to certain regularity \textit{a priori}.

The most straightforward way is a random sampling along the canonical simplex. This was first studied in some predecessors of MOEA/D. For example, MOGLS~\cite{IshibuchiM98,Jaszkiewicz02a,Jaszkiewicz02b} randomly generates a weight vector at each generation to guide a local search upon an elite solution. In~\cite{LiDDZ15}, Li et al. proposed to use evenly distributed weight vectors at the early stage of MOEA/D and then it switches to randomly generated weight vectors when the search stagnates.

Instead of a pure randomness, some principled methods from the optimal experimental design, e.g., uniform design~\cite{TanJLW12,TanJLW13,TanJ13}, have been applied for weight vector generation. Deb and Li et al.~\cite{DebJ14,LiDZK15} proposed a two-layer weight vector generation method to improve the distribution of weight vectors inside of the canonical simplex for many-objective optimization. In~\cite{JinOS01}, Jin et al. proposed a periodical variation method for two-objective optimization problems. Given a weight vector $\mathbf{w}(t)=(w_1(t),w_2(t))^T$, it changes gradually and periodically with the process of the evolution as $\{w_1(t)=|\sin(2\pi t)/F|,w_2(t)=1.0-w_1(t)\}$ where $F$ is a parameter that controls the change frequency, $t$ is the number of generation and $|\cdot|$ returns the absolute value. More recently, Blank et al.~\cite{BlankDDBS21} proposed to use the Riesz $s$-energy metric to iteratively generate a set of well-spaced weight vectors in high-dimensional spaces.


\subsubsection{Model-based adaptation methods}
\label{sec:model_method}

The basic idea of this type of method is to first proactively estimate the PF shape by learning from the evolutionary population or archive. Then, a set of weight vectors are re-sampled from this estimated PF. According to the modeling methods, related works can be classified into the following two categorizes.

The first line of research aims to change the assumption of the PF from a simplex, i.e., $\sum_{i=1}^m f_i(\mathbf{x})=1$, to a parameterized polynomial format, i.e., $\sum_{i=1}^m f^p_i(\mathbf{x})=1$. In practice, the exponential term $p$ can be estimated by solving a nonlinear least square problem~\cite{LiuGC10} or by maximizing the Hypervolume of the selected non-dominated solutions from the archive~\cite{JiangCZO11}. Since the exponential term is constantly the same at different objectives in this assumption, it is hardly applicable to problems with more complex PF shapes. To alleviate this issue, Jiang et al.~\cite{JiangZO11} proposed a curve model $f_1^p(\mathbf{x})+f_2^q(\mathbf{x})=1$ where $p\neq q$ to implement an asymmetric Pareto-adaptive scheme to fit a two-objective PF. To enable the model to estimate disconnected PF segments, DMOEA/D~\cite{GuLT12} uses piecewise hyperplanes estimated by non-dominated solutions in the archive to serve the purpose.

Different from using a fixed assumption \textit{a priori}, another line of research is based on learning a parametric model from data, i.e., the evolutionary population. For example, MOEA/D-LTD~\cite{WuKJLZ17,WuLKZZ19} uses the Gaussian process regression to model the underlying PF. It makes a general assumption about the PF as a weighted polynomial format $\sum_{i=1}^m c_if_i(\mathbf{x})^{a_i}=1$ where $a_i>0$ and $c_i>0$, $i\in\{1,\cdots,m\}$. A large amount of weight vectors are first randomly sampled from this PF model. Then dominated samples along with those having a large prediction variance are removed. In particular, this helps remove samples in the disconnected regions or beyond the PF. At the end, an iterative trimming process is implemented to repeatedly remove the weight vector having the highest density until the required number of weight vectors are identified. MOEA/D-SOM uses the self-organizing map (SOM)~\cite{Kohonen82} to learn a latent model of the PF and the topological properties are preserved through the use of neighborhood functions~\cite{GuC18}. A set of uniformly distributed weight vectors are thereafter sampled from the SOM model. Later, the growing neural gas model~\cite{Fritzke94}, a variant of SOM, is applied in a similar manner to facilitate the weight vector generation in many-objective optimization~\cite{LiuIMN20,LiuJHRY20}.

\subsubsection{Delete-and-add methods}
\label{sec:delete_and_add_method}

Different from the model-based adaptation methods, which take the global population distribution into consideration and generate the entire set of weight vector at once, the delete-and-add methods are more like local tuning. It removes the invalid weight vectors, which either lie in a region without feasible solutions or an overly crowded area, and adds new ones in promising regions. 

A typical way is to amend new weight vectors in less crowded and promising areas. As a representative MOEA/D variant with an adaptive weight adjustment, MOEA/D-AWA periodically deletes weight vectors from the overly crowded regions and it adds new ones at the sparse areas guided by the external archive~\cite{QiMLJSW14}. To improve the performance of MOEA/D-AWA for many-objective optimization problems, some variants have been proposed to use a uniformly random method to generate the initial weight vectors~\cite{FariasBBA18,FariasA19}; or emphasize more on promising yet under-exploited areas~\cite{LiY20,WangWW16}; or enhance denser alternative weight vectors close to the valid weight vectors~\cite{GeZSWTZC19,LiuCZD16,LiuCZD18}.

Instead of targeting at the spare regions, some methods attempt to generate new weight vectors away from the current promising areas. For example, in RVEA$^\ast$~\cite{ChengJOS16}, it first partitions the population into different subpopulations based on the current weight vector set. For each empty subpopulation, the associated weight vector will be replaced with a unit vector, which is randomly generated inside the range specified by the minimum and maximum objective values calculated from solutions in the current population; while the other weight vectors associated with the nonempty subpopulations remain unchanged. Likewise, some other variants replace each inactive weight vector with a new one defined by a solution least similar to the current active weight vectors where the similarity is measured by either the angle~\cite{ZhaoZZDY18,GuCLL18,LiuJHR19}, cosine value~\cite{ZhouDZLM18} or perpendicular distance~\cite{AsafuddoulaSR18}.

A few methods generate new weight vectors within a regulated neighborhood. For example, A-NSGA-III~\cite{JainD14}, A$^2$-NSGA-III~\cite{JainD13} and AMOEA/D~\cite{CamachoPBI19} periodically remove invalid weight vectors having no associated solution while they generate a simplex with $m$ neighboring weight vectors for each crowded weight vector. Other types of neighborhoods include a radius~\cite{ZhangZLCC18,ZhuCFS16}, an intermediate location between two effective neighboring weight vectors~\cite{CaiMF18}, nearest non-dominated neighboring solutions~\cite{LiS08,LiL11}, or a dichotomy/bisector of weight vector pairs between the optimal solutions of two neighboring superior weight vectors~\cite{XuZZZ19,HamadaNKO11,ShiodaO15}.

Another way to implement weight vector adjustment is to reset weight vectors based on the distribution of non-dominated solutions within certain local niches. For example, AWD-MOEA/D~\cite{GuoWW15}, FV-MOEA/D~\cite{JiangFHNOZT16} and AR-MOEA~\cite{TianCZCJ18} directly use the non-dominated solutions in the archive to reset weight vectors. Differently, MOEA-ABD~\cite{ZhangTLG18} and E-IM-MOEA~\cite{LinLJ18} periodically group non-dominated solutions into multiple subsets so as for the corresponding weight vectors. Then new weight vectors are reset with respect to each subset. To improve the performance of MOEA-ABD for many-objective optimization, MOEA/D-AWVAM resets weight vectors proportional to the search difficulty of each subset, which is roughly estimated as the number of associated non-dominated solutions~\cite{HaradaHH17}. 

\subsubsection{Discussions}
\label{sec:discussion_weights}

Given the awareness of the inability of evenly distributed weight vectors to always lead to a satisfactory distribution of solutions, weight vector adaptation has been one of most active topics in recent MOEA/D research. The fixed methods are either too stochastic to provide an effective guidance to the evolution or too restricted to be generalized to various PF shapes. Although both the model-based adaptation and the delete-and-add methods are designed to be adapt to the PF shape, they heavily depend on the effectiveness of the evolution. It is anticipated that the weight vector adaptation can be misled by a poorly converged population that is either trapped by local optima or some highly deceptively region(s) of the PF. In addition, how to generate appropriate weight vectors in a high-dimensional objective space is still an open challenge. In particular, it is highly unlikely to well represent a high-dimensional objective space with a limited number of weight vectors due to the curse-of-dimensionality. On the other hand, generating a large amount of weight vectors overly increase the computational burden in practice.

%% file: subproblems.tex

\begin{table}[t!]
  \centering
  \caption{Selected works for subproblem formulations in MOEA/D.}
  \label{tab:subproblems}
\resizebox{\textwidth}{!}{ 
    \begin{tabular}{c|c|c}
    \hline
    \textsc{Subproblem Formulation} & \textsc{Algorithm Name} & \textsc{Core Technique} \\
    \hline\hline
    \multirow{1}[4]{*}{TCH variants} & AASF~\cite{Miettinen99}, MOEA/AD~\cite{WuLKZ20} & \multirow{1}[4]{*}{Adapt the weighted metric parameter} \\
\cline{2-2}          & MOEA/D-$par$~\cite{WangZZ15}, MOEA/D-PaS~\cite{WangZZ16}, MSF and PSF~\cite{JiangYWL18} & \\
    \hline\hline
    \multirow{2}[6]{*}{PBI variants} & NSGA-III-AASF and NSGA-III-EPBI~\cite{SinghD20}, MOEA/D-PaP~\cite{MingWZZ17} & \multirow{1}[4]{*}{Adapt the penalty parameter} \\
    \cline{2-2}          & MOEA/D-APS and MOEA/D-SPS~\cite{YangJJ17} & \\
\cline{2-3}          & MOEA/D-IPBI~\cite{Sato14,Sato15} & Inverted PBI method \\
\cline{2-3}          & MOEA/D-LTD~\cite{WuLKZZ19} & Augmented multiple distance metrics \\
    \hline\hline
    \multirow{1}[4]{*}{Constrained Decomposition} & MOEA/D-ACD~\cite{WangZZGJ16}, MOEA/D-LWS~\cite{WangZILZ18} & \multirow{1}[4]{*}{Reduce the improvement region} \\
\cline{2-2}          & MOEA/D-M2M~\cite{LiuGZ14,LiuCDG17}, MOEA/D-AM2M~\cite{LiuCZD18}  & \\
    \hline
    \end{tabular}%
    }
\end{table}

\subsection{Subproblem Formulations}
\label{sec:subproblems}

Subproblem formulation determines the search behavior of MOEA/D. Most subproblem formulations used in MOEA/D are derived from the three traditional ones introduced in~\pref{sec:decomposition}. Here we are mainly interested in the shape of the contour line and the size of the improvement region that defines the region with better solutions. In particular, a larger improvement region leads to a stronger selection pressure on convergence; otherwise it stresses more on the diversity. \pref{tab:subproblems} summarizes the selected works for subproblem formulation overviewed in this section.

%
%

\subsubsection{TCH variants}
\label{sec:tch_variants}

\begin{figure}[t!]
	\centering
    \includegraphics[width=\linewidth]{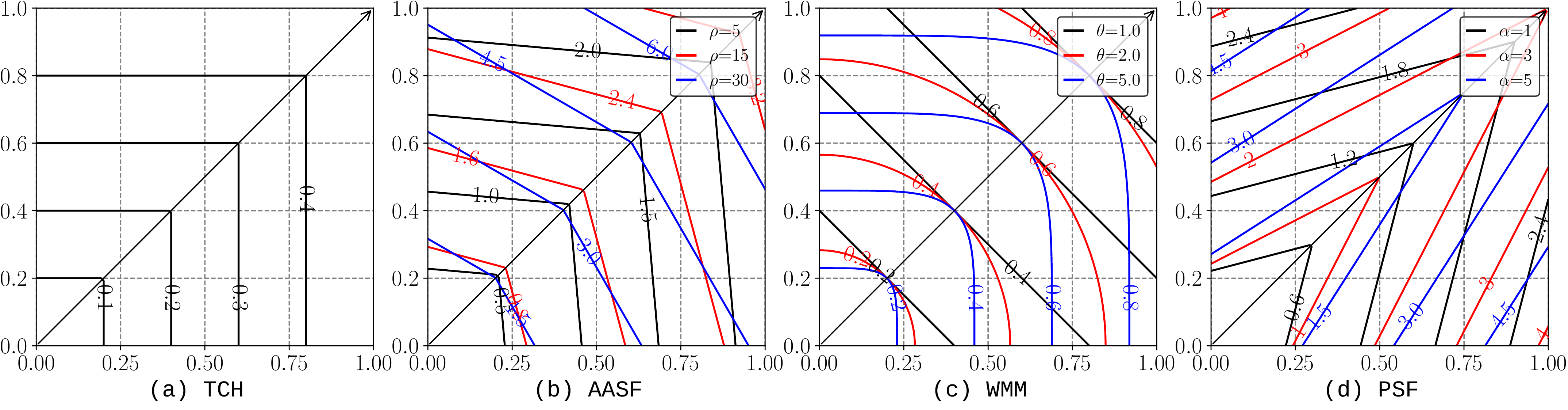}
    \caption{Contour lines of different variants of the Tchebycheff approach.}
    \label{fig:contour_tch}
\end{figure}
\vspace{-1em}

The shape of the contour line of the TCH approach and its variants mainly depends on the distance measure with respect to the ideal point. As shown in~\pref{fig:contour_tch}(a), the Tchebycheff distance used in the original TCH approach leads to a contour line resemblance to the Pareto dominance relation. It is thus not surprising that the TCH approach is asymptomatically equivalent to the Pareto dominance relation for identifying better solutions~\cite{GiagkiozisF15}. Although the TCH approach is in principle not restricted to the convexity of the PF like the WS approach, it is not able to distinguish between weakly dominated solutions. One of the most straightforward alternatives to address this issue is the use of an augmented achievement scalarizing function (AASF)~\cite{Miettinen99} 
\begin{equation}
\min_{\mathbf{x}\in\Omega}\quad g^{\mathtt{aasf}}(\mathbf{x}|\mathbf{w},\mathbf{z}^\ast)=\max_{1\leq i\leq m}\bigg\{\frac{f_i(\mathbf{x})-z_i^\ast}{w_i}\bigg\}+\rho\sum_{i=1}^m\bigg\{\frac{f_i(\mathbf{x})-z_i^\ast}{w_i}\bigg\},
\label{eq:asf}
\end{equation}
where $\rho>0$ is a parameter that tweak the opening angle of the contour line as an example shown in~\pref{fig:contour_tch}(b). Unfortunately, there is no thumb rule to set this $\rho$ value in practice to control the improvement region. Another interesting application of the AASF is in~\cite{WuLKZ20} where an adversarial decomposition approach is proposed to leverage the complementary characteristics of the AASF and the conventional PBI approaches. 

Another criticism of the TCH approach is its crispy improvement region that easily leads to a loss of population diversity. To address this issue, a natural idea is to adapt the shape of the contour line to each local niche of the corresponding subproblem. For example, in~\cite{WangZZ15,WangZZ16}, Wang et al. proposed to study the characteristics of a family of weighted metrics methods in the MCDM community:
\begin{equation}
\min_{\mathbf{x}\in\Omega}\quad g^{\mathtt{wmm}}(\mathbf{x}|\mathbf{w},\mathbf{z}^\ast)=\bigg(\sum_{i=1}^m w_i|f_i(\mathbf{x})-z_i^\ast|^\theta\bigg)^{\frac{1}{\theta}},
\end{equation}
where $\theta\geq 1$ is a parameter that controls the geometric characteristics of the contour lines of $g^{\mathtt{wmm}}(\cdot)$ as shown in~\pref{fig:contour_tch}(c). By adaptively adjusting the $\theta$ value from a predefined set $\Theta=\{1,2,3,4,5,6,7,8,9,10,\infty\}$, we can expect to tweak the most appropriate improvement region for each subproblem. Based on the similar idea, Jiang et al.~\cite{JiangYWL18} proposed a multiplicative scalarizing function (MSF) and it is defined as:
\begin{equation}
\min_{\mathbf{x}\in\Omega}\quad g^{\mathtt{msf}}(\mathbf{x}|\mathbf{w},\mathbf{z}^\ast)=\frac{[\max_{1\leq i\leq m}(\frac{1}{w_i}|f_i(\mathbf{x})-z_i^\ast|)]^{1+\alpha}}{[\min_{1\leq i\leq m}(\frac{1}{w_i}|f_i(\mathbf{x})-z_i^\ast|)]^{\alpha}},
\end{equation}
where $\alpha$ is a parameter that controls the shape of the contours of the MSF, which becomes the TCH when $\alpha=0$ while it overlaps with the corresponding weight vector when $\alpha=+\infty$. To have an adaptive control of $\alpha$,~\cite{JiangYWL18} suggested that:
\begin{equation}
\alpha^i=\beta(1-g/G_{\max})\{m\times\min_{1<j<m}(w_j^i)\},
\end{equation}
where $\alpha^i$ is the $\alpha$ value associated with the $i$-th subproblem. Meanwhile, the authors developed another TCH variant called penalty-based scalarizing function (PSF), the contour lines of which are given in~\pref{fig:contour_tch}(d), and it is defined as:
\begin{equation}
\min_{\mathbf{x}\in\Omega}\quad g^{\mathtt{psf}}(\mathbf{x}|\mathbf{w},\mathbf{z}^\ast)=\max_{1\leq i\leq m}(\frac{1}{w_i}|f_i(\mathbf{x})-z_i^\ast|)+\alpha d,
\end{equation}
where $d$ is the same as $d_2$ in the PBI approach and $\alpha$ is a parameter that controls the balance between convergence and diversity. In particular, a larger $\alpha$ value leads to a more emphasis on the diversity. In practice, the authors proposed to linearly decrease $\alpha$ with the evolution progress and it becomes zero at the end. This makes both MSF and PSF gradually degenerate to TCH. Thus, the corresponding improvement region for each subproblem is accordingly enlarged leading to a more emphasis on the convergence and a relaxation on the diversity. 

\begin{figure}[t!]
	\centering
    \includegraphics[width=\linewidth]{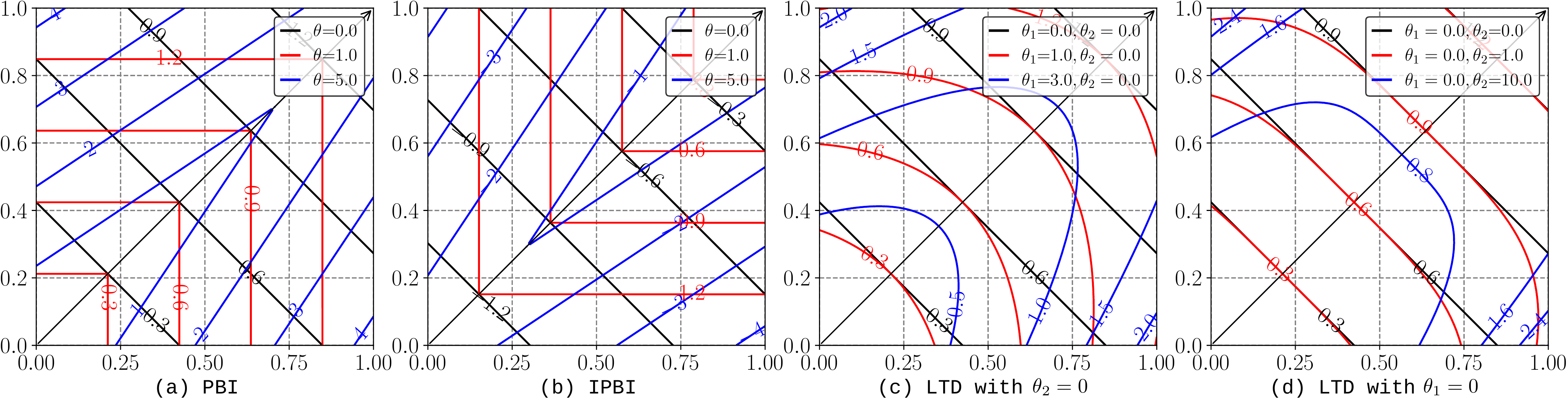}
    \caption{Contour lines of different variants of the PBI approach.}
    \label{fig:contour_pbi}
\end{figure}

\vspace{-1em}

\subsubsection{PBI variants}
\label{sec:pbi_variants}

As a parameterized combination of two distance measures, the improvement region of the PBI approach is controlled by $\theta$ in~\pref{eq:pbi}. As shown in~\pref{fig:contour_pbi}(a), it becomes the WS approach when $\theta=0.0$ and the TCH approach when $\theta=1.0$. Its improvement region becomes smaller with the increase of $\theta$. Singh and Deb~\cite{SinghD20} derived some theoretical underpins between the PBI approach with various $\theta$ values and the AASF approach. Therefore, one typical idea of the PBI variants is largely about the adaptation of $\theta$ with respect to the shape of the PF~\cite{YangJJ17,MingWZZ17}. For example, Yang et al.~\cite{YangJJ17} proposed two $\theta$ adaptation methods. One is called the adaptive penalty scheme (APS) that gradually increases the $\theta$ value with the progression of evolution:
\begin{equation}
\theta=\theta_{\min}+(\theta_{\max}-\theta_{\min})\frac{t}{t_{\max}},
\end{equation}
where $t$ is the generation count and $t_{\max}$ is the maximum number of generations. $\theta_{\min}=1.0$ and $\theta_{\max}=10.0$ are the lower and upper boundary of $\theta$, respectively. The basic idea of APS is to promote the convergence during the early stage of evolution while gradually shifting onto diversity. The alternative way, called subproblem-based penalty scheme (SPS), is to assign an independent $\theta$ value for each subproblem:
\begin{equation}
\theta_i=e^{ab_i}, b_i=\max_{1\leq j\leq m}\big\{w_i^j\big\}-\min_{1\leq j\leq m}\{w_i^j\},
\end{equation}
where $\theta_i$ means a $\theta$ value for the $i$-th subproblem in PBI and $w_i^j$ is the $j$-th element for the $i$-th weight vector. $a>0$ is a scaling factor that controls the magnitude of penalty. 

Instead of tweaking the penalty term in the PBI approach, Sato proposed an inverted PBI (IPBI)~\cite{Sato14,Sato15} as:
\begin{equation}
\min_{\mathbf{x}\in\Omega}\quad g^{\mathtt{ipbi}}(\mathbf{x}|\mathbf{w},\mathbf{z}^{\mathtt{nad}})=d_1^{\mathtt{nad}}-\theta d_2^{\mathtt{nad}},
\end{equation}
where $d_1^{\mathtt{nad}}=\|\mathbf{z}^{\mathtt{nad}}-\mathbf{F}(\mathbf{x})\mathbf{w}\|/\|\mathbf{w}\|$ and $d_2^{\mathtt{nad}}=\big\|\mathbf{z}^{\mathtt{nad}}-\big(\mathbf{F}(\mathbf{x}+d^{\mathtt{nad}}_1\frac{\mathbf{w}}{\|\mathbf{w}\|}\big)\big\|$. The basic idea of IPBI approach is to push the solution away from $\mathbf{z}^{\mathtt{nad}}$ as much as possible. From the examples shown in~\pref{fig:contour_pbi}(b), we find that the IPBI approach essentially shares similar characteristics as the conventional PBI approach but it is able to search a wider range of the objective space. In~\cite{WuLKZZ19}, Wu et al. proposed an augmented format of PBI approach as:
\begin{equation}
\min_{\mathbf{x}\in\Omega}\quad y(\mathbf{x}|\mathbf{n}^\ast,\mathbf{z}^\ast)=h(\mathbf{\overline{F}(x)}|\mathbf{n}^\ast,\mathbf{z}^\ast)=d_1+\theta_1 d_2^2+\theta_2 d_2^4,
\end{equation}
where $\theta_1>0$ and $\theta_2>0$ are parameters that control the shape and distribution of the opening angle and curvature contours and thus search behaviors of the underlying subproblem as shown in~\pref{fig:contour_pbi}. In practice, $\theta_1$ and $\theta_2$ are set according to the manifold structure of the underlying PF estimated by a Gaussian process regression model. 

\begin{figure}[t!]
	\centering
    \includegraphics[width=\linewidth]{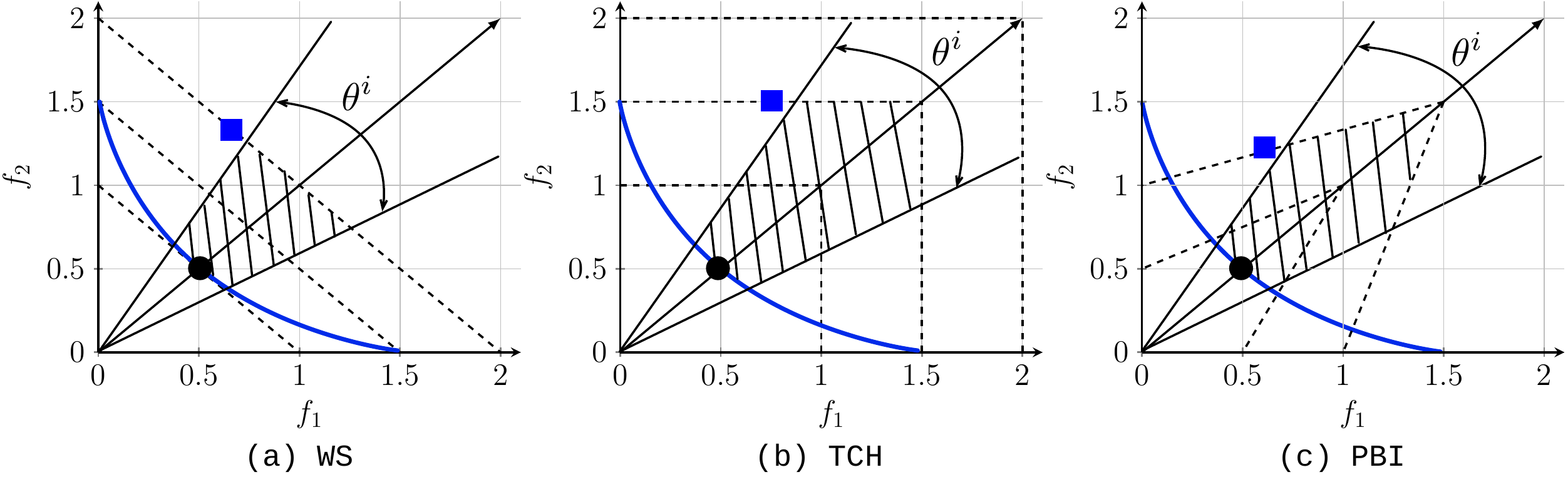}
    \caption{Contour lines of different variants CD (black circle is not comparable with the blue square due to the constrained angle $\theta^i$).}
    \label{fig:contour_cd}
\end{figure}
\vspace{-1em}

\subsubsection{Constrained Decomposition}
\label{sec:constrained_decomposition}

Instead of working on the entire search space, another decomposition method is to restrict the improvement region of each subproblem to be a local niche, also known as constrained decomposition (CD)~\cite{WangZZGJ16}. This idea is similar to some TCH and PBI variants (e.g.,~\cite{WangZZ15,WangZZ16,MingWZZ17,YangJJ17,WuLKZZ19}) that applied a dedicated parameterization on each subproblem. Specifically, a constrained optimization subproblem in CD is defined as:
\begin{equation}
\begin{array}{l l}
\min\quad\quad\quad\; g(\mathbf{x}|\mathbf{w}^i,\mathbf{z}^\ast)\\
\mathrm{subject\ to} \quad \mathbf{x}\in\Omega,\langle\mathbf{w}^i,\mathbf{F}(\mathbf{x})-\mathbf{z}^\ast \rangle\leq 0.5\theta^i
\end{array},
\end{equation}
where $g(\cdot)$ can be any subproblem formulation and $\theta^i$ controls the improvement region of the $i$-th subproblem. Based on this definition, the one that meets the constraint is superior than the violated one(s). If both solutions meet the constraint or not, they are compared based on the $g(\cdot)$ function. As the example shown in~\pref{fig:contour_cd}, the improvement region of the constrained subproblem is much smaller than its conventional version thus enabling a better population diversity. In~\cite{WangZILZ18}, a local constraint is imposed for WS to come up with a localized WS (LWS):
\begin{equation}
\begin{array}{l l}
\min\quad\quad\quad\;  g(\mathbf{x}|\mathbf{w}^i,\mathbf{z}^\ast)\\
\mathrm{subject\ to} \quad \mathbf{x}\in\Omega,\langle\mathbf{w}^i,\mathbf{F}(\mathbf{x})\rangle\leq\omega^i
\end{array},
\end{equation}
where $\omega^i=\sum_{j=1}^m\frac{\theta^{ij}}{m}$ is the constraint for the $i$-th subproblem and $\theta^{ij}$ is an acute angle between the $i$-th weight vector and its $j$-th closest weight vector. In particular, $g(\mathbf{x}^\ast|\mathbf{w}^i,\mathbf{z}^\ast)$ is set to infinity when $\langle\mathbf{w}^i,\mathbf{F}(\mathbf{x}^\ast)\rangle>\omega^i$, i.e., $\mathbf{x}^\ast$ is out of comparison when consider the $\omega^i$ constraint. Each constraint imposes a hypercone region with regard to each subproblem, similar to the effect of the example shown in~\pref{fig:contour_cd}(a). By doing so, we can expect to make solution(s) lying in the non-convex region survive in the update procedure of MOEA/D when using the LWS approach. 

Instead of decomposing the original MOP into a scalar optimization problem, Liu et al.~\cite{LiuGZ14} proposed MOEA/D-M2M that decomposes a MOP into $K>1$ simplified MOPs where the $i$-th one is defined as:
\begin{equation}
\begin{array}{l l}
\min\quad\quad\quad \mathbf{F}(\mathbf{x})=(f_1(\mathbf{x}),\cdots,f_m(\mathbf{x}))^T\\
\mathrm{subject\ to}\;\;\; \mathbf{x}\in\Omega,\mathbf{F}(\mathbf{x})\in\Delta^i\\
\end{array},
\end{equation}
where $\Delta^i=\{\mathbf{F}(\mathbf{x})\in\mathbb{R}^m|\langle\mathbf{F}(\mathbf{x}),\mathbf{w}^i\rangle\leq\langle\mathbf{F}(\mathbf{x}),\mathbf{w}^j\rangle\}$, $j\in\{1,\cdots,N\}$ and $\langle\mathbf{F}(\mathbf{x}),\mathbf{w}\rangle$ is the acute angle between $\mathbf{F}(\mathbf{x})$ and $\mathbf{w}$. In practice, each simplified MOP is solved by a dedicated NSGA-II in a collaborative manner. Since each subregion is enforced with a population of solutions, MOEA/D-M2M naturally strikes a balance between convergence and diversity. The effectiveness of this M2M decomposition approach is validated on problems with some challenging bias towards certain regions of the PF as well as those with disconnected segments~\cite{LiuCDG17} and many objectives~\cite{LiuCZD18}. 

\subsubsection{Discussions}
\label{sec:discussions_subproblem}

All existing subproblem formulations are in a parametric format, to which the search behavior is controlled by the corresponding parameters. Similar to the challenges faced by the weight vector settings, there is no thumb rule to choose the most appropriate parameters for black-box problems. Although there have been several attempts to adaptively set the parameters according to the current population, it is risky to be biased due to the premature convergence. Furthermore, it is anticipated that different subproblems have distinct difficulties. How to strategically allocate computational budgets to different subproblems according to their evolutionary status is still an open issue. Although there have been many criticisms on the ineffectiveness of those three classic subproblem formulations, it is unclear whether the highly parameterized subproblems are even more difficult to solve.

%% file: selection.tex

\begin{table}[t!]
	\centering
	\caption{Selected works for selection mechanisms in MOEA/D.}
	\label{tab:selection}
\resizebox{\textwidth}{!}{
    \begin{tabular}{c|c|c}
    \hline
    \textsc{Selection Mechanism} & \textsc{Algorithm Name} & \textsc{Core Technique} \\
    \hline\hline
    \multirow{1}[6]{*}{Mating selection} & MOEA/D-DE~\cite{LiZ09}, MOEA/D-TPN~\cite{JiangY16}, MOEA/D-MRS~\cite{LiZS19} & Neighborhood versus global \\
\cline{2-3}          & ENS-MOEA/D~\cite{ZhaoSZ12}, MOEA/D-ATO~\cite{ZhangZLZP17} & Ensemble neighborhood \\
\cline{2-3}          & MO-RCGA~\cite{GholaminezhadI14}, SRMMEA~\cite{LiSZ19}, MOEA/D-EAM~\cite{WangZLZR18} & Neighborhood of solutions \\
    \hline\hline
    \multirow{3}[10]{*}{Environmental selection} & MOEA/D-DE~\cite{LiZ09}, EFR~\cite{YuanXW14a} & Neighborhood versus global \\
\cline{2-3}          & MOEA/DD~\cite{LiDZK15}, ISC-MOEA/D~\cite{ElarbiBS17}, g-DBEA~\cite{AsafuddoulaSR18}, DDEA and DDEA+NS~\cite{AsafuddoulaSR18} & \multirow{1}[4]{*}{Prioritizing isolated regions}\\ 
\cline{2-2}          & VaEA~\cite{HeZCZ19}, OLS~\cite{ZhouCHX20}, ASEA~\cite{LiuZYERS19} & \\
\cline{2-3}          & MOEA/D-STM\cite{LiZKLW14}, AOOSTM and AMOSTM\cite{WuLKZZ17}, MOEA/D-IR~\cite{LiKZD15} & Matching solutions and subproblems \\
\cline{2-3}          & MOEA/D-ARS~\cite{ChenSZWS19}, MOEA/D-CD and MOEA/D-ACD~\cite{WangZZGJ16}, MOEA/D-GR~\cite{WangZGZ14,WangZZGJ16} & Global replacement \\
\cline{2-3}          & $\theta$-DEA~\cite{YuanXWY16}, MOEA/D-DU and EFR-RR~\cite{YuanXWZY16}, MOEA/D-CSM~\cite{LiuWBLJ20} & Local niche competition \\
    \hline
    \end{tabular}
}
  \label{tab:addlabel}
\end{table}

\vspace{-1em}
\subsection{Selection Mechanisms in MOEA/D}
\label{sec:selection}


Selection not only helps decide how to select high quality mating parents for offspring reproduction, as known as the mating selection; but also implements the survival of the fittest that controls the evolution, as known as the environmental selection. \pref{tab:selection} summarizes the related works on both mating and environmental selection mechanisms.

\subsubsection{Mating selection}
\label{sec:mating_selection}

In the original MOEA/D~\cite{ZhangL07}, the mating selection is restricted to the neighborhood of each subproblem. This is based on the assumption that neighboring subproblems share certain structural similarity thus their corresponding solutions are expected to help each other. However, this mechanism is too exploitive thus has a risk of being trapped by local optima. A simple strategy is to allow mating parents to be picked up from the entire population with a low probability~\cite{LiZ09}. Differently, MOEA/D-TPN~\cite{JiangY16} proposed to take the local crowdedness into consideration when selecting mating parents. If a neighborhood is associated with too many solutions, the mating parents will be selected outside of this neighborhood. Likewise, MOEA/D-MRS~\cite{LiZS19} proposed a self-adaptive mating restriction strategy that dynamically sets the mating selection probability according to the estimated quality of solutions associated with each subproblem. In particular, if a subproblem is frequently updated, it is assumed to be associated with a poor solution. Accordingly, it is assigned with a small mating selection probability and vice versa.

Since the mating selection is mainly implemented within the neighborhood of each subproblem, it is anticipated that the size of each neighborhood can influence the balance of exploration versus exploitation (EvE). Instead of using a constant neighborhood size for each subproblem, \cite{ZhaoSZ12,ZhangZLZP17} proposed to maintain an ensemble of neighborhood size settings which compete with each other to implement an adaptive EvE balance. Instead of using the neighborhood of each subproblem, an alternative is to use the neighborhood of solutions in mating selection, such as clusters of solutions by using $k$-means algorithm~\cite{GholaminezhadI14,LiSZ19} or solutions close to each other in the objective space~\cite{WangZLZR18}.

\subsubsection{Environmental selection}
\label{sec:environmental_selection}

The original MOEA/D uses a steady-state environmental selection mechanism that the offspring solution is directly used to update the parent population right after its generation. The offspring can replace as many parent solutions as it can within the mating neighborhood. Obviously, such greedy strategy is detrimental to the population diversity~\cite{LiZ09}. To amend this issue, it uses the same strategy as in the mating selection to give a small chance to update the subproblems outside of the mating neighborhood. To give a more robust fitness with regard to a neighborhood area, Yuan et al.~\cite{YuanXW14a} proposed an ensemble fitness ranking (EFR) method to assign fitness value to each solution. It aggregates the ranks of scalarizing function values for a group of subproblems with regard to a solution.

Different from most environmental selection mechanisms that always prioritize the survival of non-dominated solutions against dominated ones, MOEA/DD~\cite{LiDZK15} is the pioneer that suggests emphasizing the survival of solutions lying in isolated regions with few neighbors even if they are dominated or infeasible ones. The unpinning assumption is that these regions are likely to be under exploited. Therefore, they are beneficial to the population diversity and help mitigate the risk of being trapped by local optima. Later, the importance of maintaining isolated solutions irregardless of their convergence and feasibility is empirically investigated in~\cite{ElarbiBS17}. Inspired by MOEA/DD, there have been some further attempts that try to allocate certain survival chance to solutions that are relatively inferior in terms of the convergence but contribute more in terms of population diversity, e.g.,~\cite{XiangZLC17,AsafuddoulaSR18,HeZCZ19,ZhouCHX20}. In addition, MOEA/DD is an early work that aims to leverage the complementary effect of dominance- and decomposition-based selection strategies. Along this line, there have been some variants developed to exploit the benefits of combining various selection strategies under the same paradigm to achieve a better trade-off between convergence and diversity~\cite{LiuZYERS19}.

Instead of a steady-state scheme, \cite{LiZKLW14,WuLKZZ17} proposed the first generational environmental selection mechanism for MOEA/D. Its basic idea is to transform the environmental selection process as a matching problem among subproblems and solutions, i.e., subproblems aim to find their appropriate solutions and vice versa. In particular, the preference of subproblems is convergence while that of solutions is diversity. A stable matching among subproblems and solutions naturally achieves a balance between convergence and diversity. This stable matching based selection mechanism inspires a series of follow-up MOEA/D variants that take the matching relationship between subproblems and solutions into consideration. For example, \cite{LiKZD15} implemented another generational environmental selection mechanism that first builds the interrelationship among subproblems and solutions. Each solution is first attached to its most matching subproblem. Thereafter, the fittest solutions from those nonempty subproblems are chosen to survive to the next round while the remaining slot is filled with some less favorable solutions. This idea is extended as a global replacement mechanism that first finds the most matching subproblem for each offspring solution before updating the parent population~\cite{WangZGZ14,WangZZGJ16,WangZZGJ16a,ChenSZWS19}, a local niche competition~\cite{YuanXWZY16,YuanXWY16} and a correlative selection mechanism~\cite{LiuWBLJ20}.

\vspace{-1em}
\subsubsection{Discussions}
\label{sec:discussion}

Selection is one of the most critical steps in both natural evolution and EAs. In a nutshell, the major target of both mating and environmental selections is to achieve an EvE dilemma and also a balance between convergence and diversity. Nonetheless, given the black-box nature of the underlying problem and also the dynamics of evolution, there is no thumb rule to evaluate neither exploration nor exploitation (the same for the convergence and diversity). Furthermore, environmental selection is the springboard for analyzing the search dynamics of MOEA/D. However, related theoretical analysis has been significantly overlooked in the literature.

%% file: reproduction.tex

\begin{table}[t!]
	\centering
	\caption{Selected works for reproduction operators in MOEA/D.}
	\label{tab:reproduction}
\resizebox{\textwidth}{!}{
\begin{tabular}{c|c|c}
\hline
\textsc{Reproduction   Operator}          & \textsc{Algorithm Name}                                                                   & \textsc{Core Technique}                      \\ \hline\hline
\multirow{4}{*}{Swarm intelligence}       & dMOPSO {[}146{]}, SDMOPSO   {[}163{]}, MOPSO/D {[}177{]}, D$^2$MOPSO {[}164{]}            & \multirow{2}{*}{PSO}                         \\
                                          & MMOPSO {[}125{]}, DEMPSO {[}170{]}, MS-GPSO/D$^\text{w}$ {[}253{]}, MaPSO{[}229{]}               &                                              \\
\cline{2-3}                                          & MOEA/D-ACO {[}95{]}, ACO-Index {[}250{]},  NMOACO/D {[}168{]}, MACO/D-LOA   {[}228{]}     & ACO                                          \\
\cline{2-3}                                          & dMOABC {[}254{]}, MOCS/D {[}20{]},  MOGWO {[}158{]}, MIWOALS {[}1{]}                      & Other SI                                     \\ \hline\hline
\multirow{6}{*}{Model-based methods}      & MEDA/D {[}47, 255{]}                                                                      & EDAs                                         \\
\cline{2-3}                                          & MOEA/D-GM {[}30{]}                                                                        & Probabilistic graphic model                  \\
\cline{2-3}                                          & MOEA/D-PBIL {[}233{]}                                                                     & PBIL      \\
\cline{2-3}                                          & MACE-gD {[}53{]}                                                                          & Cross-entropy method                         \\
\cline{2-3}                                          & MOEA/D-CMA {[}151{]}, MO-CMA-D {[}172{]},MOEA/D-CMA {[}112{]}                             & CMA-ES                                       \\
\cline{2-3}                                          & MOPC/D {[}161{]}                                                                          & Probability   collectives                    \\ \hline\hline
\multirow{6}{*}{AOS and HH}               & MOEA/D-FRRMAB {[}117{]},   MOEA/D-UCB-Tuned and MOEA/D-UCB-V {[}55{]}                     & \multirow{2}{*}{Adaptive operator selection} \\
                                          & NSGA-III$_{\text{AP}}$ and NSGA-III$_{\text{PM}}$ {[}58{]}, SaMOEA/D {[}184{]}, MOEA/D-DYTS {[}198{]} &                                              \\
\cline{2-3}                                          & MPADE {[}25{]}, MOEA/D-CDE {[}126{]}, AMOEA/D {[}127{]}                                   & Parameter control                            \\
\cline{2-3}                                          & MOEA/D-GHDE {[}140{]}, EF-PD {[}218{]}, MOEA/D-IMA {[}231{]}                              & Ensemble of operators                        \\
\cline{2-3}                                          & MAB-HH {[}45{]}, MOEA/D-HH\_\{SW\} {[}56{]}, MOEA/D-HH {[}57{]}, MOEA/D-DRA-UCB {[}181{]} & \multirow{2}{*}{Hyper-heuristics}            \\
                                          & BoostMOEA/D-DRA {[}182{]}, MOEA/D-LinUCB {[}54{]}, MAB-based HH {[}27{]}                  &                                              \\ \hline\hline
\multirow{3}{*}{Mathematical programming} & MOEA/D-LS {[}147{]},   MOEA/D-RBF {[}149{]}, MONSS {[}150{]}, NSS-MO {[}152{]}            & \multirow{2}{*}{Direct search method}        \\
                                          & MOEA/D-QS{[}180{]}, MOEA/D-NMS {[}245{]}                                                  &                                              \\
\cline{2-3}                                          & NSGA-III-KKTPM {[}2, 3, 191{]}                                                            & KKTPM                                        \\ \hline\hline
\multirow{5}{*}{Memetic search}           & MOMAD {[}96{]}, PPLS/D   {[}193{]}                                                        & Pareto local search                          \\
\cline{2-3}                                          & CoMOLS/D {[}15{]}, MONSD {[}73{]},    MOMNS/D and MOMNS/V {[}211{]}, LS/D {[}258{]}       & \multirow{2}{*}{Neighborhood search}         \\
                                          & MOEA/D-TS {[}5{]}, MOEA-MA {[}38{]}, DMTS {[}259{]}, MOEA/D-GLS {[}4{]}                   &                                              \\
\cline{2-3}                                          & EMOSA {[}109,110{]}, MOMA-SA {[}14{]}, EOMO {[}90{]}                                      & \multirow{2}{*}{Simulated annealing}         \\
                                          & MOEA/D-GRASP {[}6{]}, HEMH {[}94{]}                                                       &                                              \\

\hline              
\end{tabular}
}
\end{table}

\vspace{-1em}
\subsection{Reproduction Operators in MOEA/D}
\label{sec:reproduction}

Reproduction operators are used to generate offspring solutions thus determine the way to explore the search space. The original MOEA/D~\cite{ZhangL07} applied the simulated binary crossover and polynomial mutation for offspring reproduction whereas the differential evolution considered in~\cite{LiZ09} has a much wider uptake thereafter. In principle, any reproduction operator proposed in the EA community can be used in MOEA/D. Here we overview some representative operators being used in MOEA/D according to their types, as summarized in~\pref{tab:reproduction}.

\vspace{-.3em}
\subsubsection{Swarm intelligence methods}
\label{sec:swarm}

Swarm intelligence (SI) is the collective intelligence behavior of self-organized and decentralized systems such as artificial groups of simple agents. Examples of SI include the group foraging of social insects, cooperative transportation, nest-building of social insects, and collective sorting and clustering.
\begin{itemize}
	\item\underline{\textit{Particle swarm optimization (PSO)}}: This is one of the most popular SI initially exploited in~\cite{PengZ08,MoubayedPM10,MartinezC11} under the MOEA/D framework. Like the other PSO variants, they maintain a swarm of particles, each of which carries a unique weight vector and has a position vector, a personal best vector and a velocity vector. Instead of having only one global best vector in the vanilla PSO, a population of global best vectors, each of which corresponds to a subproblem, are maintained in MOEA/D. The MOEA/D variants based on PSO mainly differ in the way of fitness assessment of each particle. For example, D$^2$MOPSO~\cite{MoubayedPM14} uses the PBI scalarizing function is used to update the personal best vector and dominance plays a major role in building the leaders' archive. \cite{LinLDCM15,Pan0GW18,YuCGZYKZ18,XiangZCZ20} proposed to use a scalarizing function defined by a weight vector to evaluate the quality of each particle.
	\item\underline{\textit{Ant colony optimization (ACO)}}: MOEA/D-ACO is the first ACO implementation under the MOEA/D framework~\cite{KeZB13} where each ant tackles a subproblem. The neighborhood is formed by dividing all ants into a few groups, each of which maintains a pheromone matrix. During the search process, each ant has a heuristic information matrix and keeps a record of the best solution found so far for its subproblem. A new candidate solution is constructed by combining information from the group pheromone matrix, the own heuristic information matrix and the current solution. Its survival follows the MOEA/D that compares the fitness with other solutions within an ant group. Further developments of MOEA/D-ACO include using opposition-based learning~\cite{ZhangXLLS20}, negative pheromone~\cite{NingZSF20} and a local optimum avoidance strategy~\cite{WuQJZX20}.
\end{itemize}

In addition to PSO and ACO, any SI method such as artificial bee colony~\cite{ZhongXL14}, cuckoo search~\cite{ChenGLXCF19}, grey wolf optimization~\cite{MirjaliliSMC16} and whale optimization~\cite{Abdel-BassetMM21} can be adapted to the MOEA/D framework. 

\subsubsection{Model-based methods}
\label{sec:model}

Traditional reproduction operators in EAs, ranging from genetic operators such as crossover and mutation to SI methods, are developed on the basis of some fixed heuristic rules or strategies. They mainly work with individual solutions yet rarely interact with the environment. It is natural that the environment varies rapidly with the progression of evolution. Furthermore, the intrinsic complexity of variable dependency renders the problem landscape twisted or rotated. In this case, it is anticipated that those traditional reproduction operators may not work effectively due to the incapacity of learning from the environment. To address this issue, there have been a number of efforts dedicated to equipping the reproduction operators with a learning ability. The basic idea is to replace the heuristic operators with machine learning models, from which new candidates are sampled.
\begin{itemize}
	\item\underline{\textit{Estimation of distribution algorithm (EDA)}}: The basic idea of EDA is to estimate the distribution of promising candidate solutions by training and sampling models in the decision space. MEDA/D is the first to exploit the multi-objective EDA based on decomposition~\cite{GaoZZ12,ZhouGZ13} for multi-objective traveling sales man problem (MOTSP). Specifically, MEDA/D decomposes a MOTSP into a set of subproblems, each of which is with a probabilistic model. Each model consists of both priori heuristic information and learned information from the evolving population. New candidate solutions, i.e., new tours, are directly sampled from those probabilistic models. The approximation of the PF of a MOTSP is implemented by the cooperation of neighboring subproblems under the MOEA/D framework. Other more expressive probabilistic models (e.g., probabilistic graphic model~\cite{desouza2015moeadgm}, population-based incremental learning (PBIL)~\cite{XingWLLQ17}, cross-entropy method~\cite{GiagkiozisPF14}, and probability collectives\cite{MorganWC13}) and a hybrid strategy thats combine EDA and other reproduction operators have also been studied.
	
	\item\underline{\textit{Covariance matrix adaptation evolution strategy (CMA-ES)}}: The initial attempts on incorporating CMA-ES into MOEA/D were in~\cite{ParkL13,MartinezDLBAT15}. Since they associate each subproblem with a unique Gaussian model, their efficiency exponentially decreases with the increase of the population size. To address this issue, MOEA/D-CMA~\cite{LiZD17} considers a hybrid strategy between differential evolution and CMA-ES. Specifically, each Gaussian model is responsible for a group of subproblems, in which only one subproblem is optimized at a time. If a CMA-ES converges or the corresponding Gaussian model becomes ill-conditioned, it picks up another subproblem from the same group to optimize.	
\end{itemize}



\subsubsection{Adaptive operator selection and hyper-heuristics}
\label{sec:aos}

It is widely appreciated that different reproduction operators have their distinctive characteristics and search dynamics. Any reproduction operator is arguable to be versatile across various kinds of problems. The EvE dilemma is expected to dynamically change with the problem-solving progress as it requires more exploration at the beginning while it turns down to be more exploitive when approaching the optimum.

Based on this justification, instead of constantly using one reproduction operator, it is more attractive to automatically choose the most appropriate one according to the feedback collected during the search process, as known as adaptive operator selection (AOS). Although AOS have been extensively studied in single-objective optimization, it is yet developed in the context of multi-objective optimization until MOEA/D-FRRMAB~\cite{LiFKZ14}. This is largely attributed to the challenge of fitness assignment when encountering multiple conflicting objectives. Due to the use of subproblems, MOEA/D provides a natural springboard for implementing AOS in multi-objective optimization. As the first AOS paradigm in MOEA/D, FRRMAB~\cite{LiFKZ14} consists of two functional components, i.e., credit assignment and operator selection. The credit assignment component is used to measure the effectiveness of using a reproduction operator. The FRRMAB uses the fitness improvement rate with respect to each subproblem as the building block for this purpose. Furthermore, it also applies a sliding window to prioritize more recent observations. As for the operator selection, which decides the next reproduction operator for generating new candidate, the FRRMAB transforms the AOS as a multi-armed bandit problem and applies the upper confidence bound (UCB) algorithm in practice. Based on the FRRMAB, many other AOS variants have been developed afterwards, such as~\cite{GoncalvesAP15,QiBMML16,GoncalvesPAKVD17,SunL20}. In addition to AOS, \cite{LinLYDCLWC16,CuiLLCL16,LinTMDLCM17} also consider adaptively control the parameters associated with the reproduction operator in order to achieve the best algorithm setup. Rather than adaptively selecting the \lq\lq appropriate\rq\rq\ reproduction operator on the fly, another line of research (e.g.,~\cite{WangYLZWCC19,XieQWY20,LuoLJJC21}) is to build an ensemble of reproduction operators and use them simultaneously.

Hyper-heuristics (HH) is a high-level methodology for selection or automatic generation of heuristics for solving complex problems. Different from AOS, HH does not just pick up reproduction operators from the existing portfolio, it has the ability to create new operators co-evolving with the problem-solving process. \cite{GoncalvesKAV15a,GoncalvesKAV15b,PrestesDGAP17,FerreiraGP17} were the first attempts of using HH under the MOEA/D framework where a FRRMAB variant was developed to determine the low-level heuristic generation or selection. Besides, HH is combined with other heuristics such as irace~\cite{PrestesDLGA18} and bandit models~\cite{GoncalvesALD18,AlmeidaGVLD20} to composite two levels of heuristics.

\subsubsection{Mathematical programming methods}
\label{sec:mathematical}

Although reproduction operators introduced in the previous subsections are assumed to be self-adaptive in balancing EvE dilemma with evolution, they are usually more prone to excel in a global search in practice. To improve the exploitation upon certain promising areas, there have been some attempts on integrating mathematical programming methods, based on either gradient information or being derivative-free, with MOEA/D as local search operators, also known as hybrid methods. The Nelder and Mead's method~\cite{NelderM65}, a derivative-free optimization method that implements a nonlinear simplex search by alternating among three movements, i.e., reflection, expansion and contraction, performed in the simplex, is one of the most popular direct search methods to be applied in MOEA/D~\cite{MartinezMC11,MartinezC12,MartinezC13,MartinezC16,PrestesAG15,ZhangZZS17}. Recently, Deb and Abouhawwash proposed a proximity measure based on the Karush-Kuhn-Tucker optimality theory~\cite{DebA16}, dubbed KKTPM. This measure was originally designed to evaluate the convergence of a set of non-dominated solutions with regard to the PS. Later, it has also been used as either a driving force or a termination criterion of a local search procedure in NSGA-III~\cite{DebAS17,AbouhawwashSD17,SeadaAD19}. 

\subsubsection{Memetic search methods}
\label{sec:memetic}

Memetic algorithm~\cite{MoscatoN92} is an extension of the traditional genetic algorithm that uses separate individual learning or local search techniques to mitigate the likelihood of the premature convergence. In the context of MOEA/D, memetic algorithms are mainly proposed for solving combinatorial MOPs as opposed to the local search based on mathematical programming approaches in~\pref{sec:mathematical}. Here we overview some representative developments of memetic MOEA/D according to the type of local search.
\begin{itemize}
	\item\underline{\textit{Pareto local search (PLS)}}: Its basic idea is to explore the neighborhood of solutions to find either new non-weakly dominated neighborhoods~\cite{LustT10} or a better local optimum with regard to a weighted sum aggregation~\cite{PaqueteS03}. MOMAD~\cite{KeZB14} is the first one that integrates PLS into the MOEA/D framework where the PLS is conducted upon an auxiliary population. In view of the high computational cost of PLS for obtaining a good approximation of the PF, Shi et al.~\cite{ShiZS20} proposed to use parallel computation and problem decomposition to accelerate PLS. 
	
	\item\underline{\textit{Neighborhood search (NS)}}: This local search technique aims to find good or near-optimal solutions by repeatedly exploiting the neighborhood of a current solution. The neighborhood constitutes a set of similar solutions perturbed by the original one. NS operators are usually applied in a synergistic manner by simultaneously using more than one operator with different neighborhood sizes or search behavior, e.g.,~\cite{HuWLWLY19,CaiHGGZFH19,ZhouKCWLHW20,WangYZGSZ21}. As NS variant, tabu search, which uses attributive memory structures to guide the search away from local optima, has also been used as a complementary operator in MOEA/D~\cite{AlhindiZ14,ZhouWWW18,DuXZCH19,AlhindiAAATA19}.
	
	\item\underline{\textit{Simulated annealing (SA)}}: SA is a local search technique originated from statistical physics~\cite{KirkpatrickGV83}. SA has been widely recognized to be effective for combinatorial optimization problems and it was first combined with MOEA/D by Li and Silva~\cite{LiS08,LiL11}. Later, the same idea was applied for solving software next release problems~\cite{CaiCFGW17} and hybrid flow shop problems~\cite{JiangZ19a}. In addition, 
	
	\item\underline{\textit{Greedy randomized adaptive search procedure (GRASP)}}: This is a randomized local search technique that has been applied in the MOEA/D framework in~\cite{KafafyBB11,AlhindiZT14}. 
\end{itemize}

\subsubsection{Discussions}
\label{sec:discussions}

Although many reproduction operators have been applied in MOEA/D, there does not exist one winner take all in view of the no-free-lunch theorem. In other words, a reproduction operator might only excel in one type of problems or the heuristics even need to be tailored according to the problem characteristics. Although both AOS and HH are expected to autonomously identify the bespoke operator, they are essentially an online optimization problem, which is challenging to solve on top of the underlying MOP.

%% file: advanced.tex

\section{Advanced Topics}
\label{sec:advanced}

This section plans to overview some important developments of MOEA/D variants for some specific scenarios.

\begin{table}[t!]
	\centering
	\caption{Selected works for constraint handling in MOEA/D.}
	\label{tab:constraint}
\resizebox{\textwidth}{!}{
    \begin{tabular}{c|c|c}
    \hline
    \textsc{Subcategory} & \textsc{Algorithm Name} & \textsc{Core Technique} \\
    \hline\hline
    \multirow{1}[6]{*}{Feasibility-driven method} & CMOEA/D-DE-ATP~\cite{JanZ10,JanK13}, g-DBEA~\cite{AsafuddoulaRSA12,AsafuddoulaRS15} & Constraint violation \\
\cline{2-3}          & C-MOEA/D and C-NSGA-III~\cite{JainD14}, C-MOEA/DD~\cite{LiDZK15} & Prioritize isolated regions \\
\cline{2-3}          & MOEA/D-IEpsilon~\cite{FanLCHFYMWG19}, $\epsilon$-MOEA/D-$\delta$~\cite{MartinezP20}, MOEA/D-WEO-$\epsilon$CHT~\cite{WangZC20} & $\epsilon$-constraint \\
	\hline\hline
    Weight vector adaptation & RVEA~\cite{ChengJOS16}, g-DEBA~\cite{AsafuddoulaSR18}, CM2M~\cite{PengLG17} & Weight vector adaptation \\
    \hline\hline
    \multirow{1}[4]{*}{Trade-off among convergence, diversity and feasibility} & C-TAEA~\cite{LiCFY19}, MOEA/D-TW~\cite{Zhu0LS19}, IDW-M2M-CDP~\cite{PengLG20} & Multi-population \\
\cline{2-3}          & PPS-MOEA/D~\cite{FanLCLWZDG19}, PPS-M2M~\cite{FanWLYYYSR20}, DC-NSGA-III~\cite{JiaoZLYO20} & Push and pull \\
    \hline
    \end{tabular}
}
\end{table}%

\vspace{-1em}
\subsection{Constraint Handling in MOEA/D}
\label{sec:constraint}

Most, if not all, real-life optimization scenarios have various constraints by nature. However, constraint handling has not been discussed in the seminal paper of MOEA/D~\cite{ZhangL07}. Generally speaking, the direct impact of constraints is the infeasible region(s) in the search space that lead to obstacles to an effective evolutionary search. In addition, some constraints can also lead to a distortion of the PF in terms of its shape and integrity. The ideas of the existing constraint handling techniques in MOEA/D can be divided into the following three categories, as summarized in~\pref{tab:constraint}.

\vspace{-.5em}
\subsubsection{Feasibility-driven methods}
\label{sec:feasibility}

The first category is mainly driven by the feasibility information where feasible solutions are always prioritized to survive. CMOEA/D-DE-ATP, which directly aggregates penalty function into the scalarizing function, is the first attempt of considering constraint handling in MOEA/D~\cite{JanZ10}. By adaptively tweaking the violation threshold according to the type of constraints, size of the feasible region and the search outcome, g-DBEA~\cite{AsafuddoulaRSA12,AsafuddoulaRS15} proposed an improved constraint violation calculation method in MOEA/D. The constraint-domination principle, which has been successful for constraint handling in NSGA-II~\cite{DebAPM02}, can also be incorporated in MOEA/D, e.g., \cite{JanK13,JainD14}. As an extension of MOEA/DD, C-MOEA/DD~\cite{LiDZK15} proposed to prioritize an infeasible solution over feasible solutions in case it is in an isolated subregion. After some preliminary encouraging results reported in~\cite{YangCF14}, variants of the classic $\epsilon$-constraint handling method have been studied in the context of MOEA/D, such as~\cite{FanLCHFYMWG19} that dynamically adjusts the $\epsilon$ level according to the ratio of the feasible solutions in the current population and~\cite{WangZC20,MartinezP20}. 

\subsubsection{Weight vector adaptation}
\label{sec:adatpation_constraint}

The second category mainly aims to adjust the distribution of weight vectors in order to adapt to the distortion of the PF caused by constraints. Its basic idea is to progressively detect the infeasible part(s) in the objective space thus to remove the weight vectors targeting those parts while supplement new ones in the feasible part(s), e.g.,~\cite{ChengJOS16,AsafuddoulaSR18}. In~\cite{PengLG17}, Peng et al. proposed to maintain two types of weight vectors to probe both feasible and infeasible regions. The infeasible weights are used to maintain a set of well distributed infeasible solutions thus to enable a better utilization of useful information. Meanwhile, they are dynamically changed along with the evolution to prefer infeasible individuals with better objective values and smaller constraint violations. 

\subsubsection{Trade-off among convergence, diversity and feasibility}
\label{sec:tradeoff}

This is an emerging strategy that tries to balancing the trade-off among convergence, diversity and feasibility. Most, if not all, currently prevalent constraint handling techniques first tend to push a population towards the feasible region as much as possible, before considering the balance between convergence and diversity within the feasible region. This might lead to the population being stuck at some local optima or local feasible regions, especially when the feasible regions are narrow and/or disparately distributed in the search space. As the first work that explicitly take this three-way trade-off into algorithm design, C-TAEA~\cite{LiCFY19,ShanL21} use two complementary and co-evolving archives to solve constrained MOPs. Specifically, the convergence archive aims to push the population towards the PF; while the diversity archive plays as an auxiliary to explore areas under-exploited by the convergence archive including the infeasible regions. Similar idea was thereafter explored in some follow-up works such as~\cite{Zhu0LS19} and~\cite{PengLG20}. In~\cite{FanLCLWZDG19,FanWLYYYSR20}, Fan et al. proposed a push and pull search (PPS) framework that bears a similar principle as~\cite{LiCFY19} but divides the search process into two independent search stages dubbed \textit{push} and \textit{pull}. In the push stage, the algorithm aims to explore the search space without considering any constraints; while in the push stage, constraint handling approaches such as $\epsilon$-constraint handling can be applied to probe and estimate the landscape of CMOPs. In~\cite{JiaoZLYO20}, Jiao et al. proposed decomposition-based constrained multi-objective evolutionary algorithm with two types of weight vectors, emphasizing respectively  convergence and diversity. The solutions associated to the convergence weight vectors are updated only considering the aggregation function in order to search the whole search space freely, while the ones associated to the diversity weight vectors are renewed by considering both the aggregation function and the overall constraint violation, which encourages to search around the feasible region found so far.

\begin{table}[t!]
	\centering
	\caption{Selected works for surrogate-assisted MOEA/D for expensive optimization problems.}
	\label{tab:surrogate}
\resizebox{\textwidth}{!}{
    \begin{tabular}{c|c|c}
    \hline
    \textsc{Subcategory} & \textsc{Algorithm Name} & \textsc{Core Technique} \\
    \hline\hline
    \multirow{2}[12]{*}{Surrogate-assisted MOEA/D} & ParEGO~\cite{Knowles06}, MOEA/D-EGO~\cite{ZhangLTV10} & EGO \\
\cline{2-3}          & MOEA/D-RBF~\cite{MartinezC13}, MOEA/D-RBF+LS~\cite{MartinezCZ13}, ELMOEA/D-DE~\cite{PavelskiDAGV14,PavelskiDAGV16}, K-CSEA~\cite{HeCJ019} & \multirow{1}[4]{*}{Other models} \\
\cline{2-2}          & MOEA/D-S$^3$~\cite{SonodaN20}, S-CMO~\cite{PruvostDLVZ20} &  \\
\cline{2-3}          & EGO-MO~\cite{FengZZTYM15}, AB-MOEA~\cite{WangJSO20}, EPBII and EIPBII~\cite{NamuraSO17}, TIC-SMEA~\cite{LiGGSH21}, HSMEA~\cite{HabibSCRM19} & New infill criterion \\
\cline{2-3}          & SA-RVEA-PCA~\cite{ZhaoZCZYSHY20}, K-RVEA(FS)~\cite{TanW20} & Dimension reduction \\
\cline{2-3}          & TEMO-MPS~\cite{MinOGG19}, MS-RV~\cite{YangDJC20}, GCS-MOE~\cite{LuoGOW19} & Transfer learning \\
    \hline
    \end{tabular}
}
\end{table}%

\vspace{-1em}
\subsection{Surrogate-Assisted MOEA/D for Expensive Optimization Problems}
\label{sec:expensive}

Many real-world problems involve time- or resource-intensive physical experiments or numerical simulations. For example, a single function evaluation based on computational fluid dynamic (CFD) simulations could take from minutes to hours. Traditional EAs, which iteratively call for a large amount of objective function evaluations, thus become less practically feasible for those computationally expensive tasks. To amend this issue, surrogate models~\cite{WangYLK21} have been widely used in EAs to play as a proxy of expensive objective functions, as summarized in~\pref{tab:surrogate}.

ParEGO~\cite{Knowles06} is the first one that introduces surrogate modeling into the decomposition-based EMO algorithms, even earlier than the canonical MOEA/D~\cite{ZhangL07}. It uses an augmented Tchebycheff function to aggregate a MOP into a scalar function by randomly sampling a normalized weight vector during each iteration. Thereafter, the efficient global optimization (EGO)~\cite{JonesSW98}, in which the expensive objective function is modeled by a Gaussian process (GP) model and the search is driven by maximizing the expected improvement with respect to the GP model, is extended to solve expensive MOPs. Four years later, MOEA/D-EGO, i.e., the MOEA/D version's EGO, was proposed by Zhang et al~\cite{ZhangLTV10}. Since MOEA/D co-evolves a population of subproblems, it will be computationally infeasible if MOEA/D-EGO builds a surrogate model for each subproblem. Instead, it uses $m$ GPs to model each of those $m$ objective functions separately. To carry out the EGO upon a subproblem, it derives the formats of the mean and the variance with respect to both weighted sum and weighted Tchebycheff functions. In~\cite{ChughJMHS18}, K-RVEA was proposed for computationally expensive many-objective optimization problems. By making use of the uncertainty estimated by the Kriging models, the adaptive distribution of weight vectors as well as the location of solutions in the objective space, K-RVEA strikes a balance between convergence and diversity in model management. To cap the computational cost for model training without significantly compromising the model accuracy, K-RVEA developed a dedicated strategy for choosing training samples.

There have been various variants derived from either ParEGO, MOEA/D-EGO or K-RVEA. Some of them replace the GP model by other machine learning models, such as radial basis function networks~\cite{MartinezCZ13,MartinezC13}, extreme learning machine~\cite{PavelskiDAGV14,PavelskiDAGV16}, multi-level surrogate models~\cite{ZhangLHJ17}, classifier~\cite{HeCJ019}, support vector machine~\cite{SonodaN20} and Walsh function~\cite{PruvostDLVZ20}. In addition to the popular maximization of the expected improvement, some efforts have been made in new infill criteria, such as a bi-objective formulation~\cite{FengZZTYM15} or a dynamic weighted aggregation of the predicted mean and variance~\cite{WangJSO20}, a closed form expected improvement of the PBI/IPBI function~\cite{NamuraSO17}, complementary infill criteria~\cite{LiGGSH21} and multiple surrogate models~\cite{HabibSCRM19}. Some efforts have been made on using dimensionality reduction techniques (e.g., principle component analysis~\cite{ZhaoZCZYSHY20} and feature selection~\cite{TanW20}) to fight against the curse of dimensionality. Recently, there have been some efforts on using the emergent transfer learning techniques to  implement a knowledge sharing and reuse under the framework of MOEA/D, e.g.,~\cite{MinOGG19,YangDJC20}. By leveraging the piecewise linear assumption in MOEA/D thus the similarity among neighboring subproblems, Luo et al.~\cite{LuoGOW19} proposed to use multi-task GP to build collaborative surrogate models for a couple of representative subproblems simultaneously. 

\begin{table}[t!]
	\centering
	\caption{Selected works for preference incorporation in MOEA/D.}
	\label{tab:preference}
\resizebox{\textwidth}{!}{
    \begin{tabular}{c|c|c}
    \hline
    \textsc{Subcategory} & \textsc{Algorithm Name} & \textsc{Core Technique} \\
    \hline\hline
    \multirow{1}[6]{*}{A priori} & MOEA/D-NUMS~\cite{LiCMY18}, R-MEAD~\cite{MohammadiOL12}, R-MEAD2~\cite{MohammadiOLD14}, IRMEAD~\cite{ZhuHJ16} & Weight vector adaptation \\
\cline{2-3}          & a-PICEA-g~\cite{WangPF13}, cwMOEA/D~\cite{PilatN15} & Co-evolution \\
\cline{2-3}          & MOEA/D-PRE~\cite{YuZSL16} & Preference region \\
	\hline\hline
    Interactive & PMA~\cite{Jaszkiewicz04}, iMOEA/D~\cite{GongLZJZ11}, MOEA/D-PLVF and NSGA-III-PLVF~\cite{LiCSY19}, IEMO/D~\cite{TomczykK20} & Progressive learning \\
    \hline
    \end{tabular}
}
\end{table}%

\subsection{Preference Incorporation in MOEA/D}
\label{sec:preference}

The MOEA/D variants reviewed so far were designed to approximate the entire PF. As discussed in~\cite{LiLDMY20}, presenting the decision maker (DM) with a set of widely spread trade-off alternatives not only increases the DM's workload, but also incurs much irrelevant or even noisy information in decision-making. In the recent decade, there is a growing trend devoted to the synergy of EMO and MCDM to incorporate the DM's preference into the evolutionary search process~\cite{WangOJ17,XinCCIHL18,LiLDMY20,LaiL021,LiLLM21}, thus to approximate the region of interest (ROI) which is usually a partial region of the PF. Since the subproblems in MOEA/D determine the location of its approximated Pareto-optimal solutions, incorporating the DM's preference into MOEA/D is thus to adjust the distribution of weight vectors according to the DM supplied preference information. As summarized in~\pref{tab:preference}, we will overview some existing developments of preference incorporation in MOEA/D according to the preference elicitation manner, i.e., either \textit{a priori} or \textit{interactively}.

\subsubsection{A priori preference elicitation}
\label{sec:a_priori}

The basic idea is to turn the originally evenly distributed weight vectors into a biased distribution according to the DM \textit{a priori} supplied preference information. Thereafter, these biased weight vectors are thus used in MOEA/D to guide the population towards the ROI. For example, \cite{LiCMY18} proposed a closed form method, dubbed NUMS, to map the originally evenly distributed weight vectors on a canonical simplex to new positions close to the reference point supplied by the DM. Different from~\cite{LiCMY18} where the biased weight vectors are set at the outset of the evolution,~\cite{MohammadiOL12,MohammadiOLD14,ZhuHJ16} proposed to gradually adjust the weight vectors towards the ROI with the evolving population. The solutions closer to the DM supplied reference point are considered to be more preferable. Accordingly, these solutions survive in the environmental selection and are normalized to turn into the corresponding weight vectors. a-PICEA-g~\cite{WangPF13} and cwMOEA/D~\cite{PilatN15} were proposed to co-evolve both the population and the weight vectors together to approximate the ROI concurrently. In particular, a-PICEA-g randomly generates a population of weight vectors during each iteration while cwMOEA/D applies a mutation operator to reproduce new candidate weight vectors. The other weight vector generation method is through a preference region usually constituted by the DM supplied reference point along with a prescribed region (e.g., a tolerence angle~\cite{MeneghiniG17} or a radius~\cite{MaLQLJDWDHZW16}) with which the new weight vectors are sampled. Differently, Yu et al.~\cite{YuZSL16} proposed to use a scaled simplex to specify the preference region. Specifically, the biased weight vectors are first initialized as the corner of the simplex while it applies a dichotomy to iteratively add the midpoints of each pair of weight vectors until the required number of weight vectors are generated.

\subsubsection{Interactive preference elicitation}
\label{sec:interactive}

Different from the \textit{a priori} methods, \textit{interactive} preference elicitation enables the DM to progressively learn and understand the characteristics of the MOP at hand and adjust her/his elicited preference information. PMA is the earliest work along this line~\cite{Jaszkiewicz04} that periodically asks the DM to rank pairs of solutions to progressively construct a set of linear value functions, each of which is a WS. Accordingly, the learned value functions are used to guide the environmental selection process in PMA thus to approximate the ROI. In~\cite{GongLZJZ11}, iMOEA/D was proposed to periodically ask the DM to pick up the most promising solutions through a \textit{virtual interaction}. These solutions are thus used to wrap up the utility function in iMOEA/D and to guide the adjustment of weight vectors. In~\cite{LiCSY19}, Li et al. developed an interactive framework for the decomposition-based EMO algorithm to find DM preferred solutions. It consists of three modules, i.e., consultation, preference elicitation, and optimization. Specifically, after every several generations, the DM is asked to score a few candidate solutions in a consultation session. Thereafter, an approximated value function, which models the DM's preference information, is progressively learned from the DM's behavior. In the preference elicitation session, the preference information learned in the consultation module is translated into the form that can be used in a decomposition-based EMO algorithm, i.e., a set of reference points that are biased toward the ROI. Recently, IEMO/D~\cite{TomczykK20} was proposed to progressively learn the DM's preference information by inviting the DM to compare pairs of candidate solutions. It applies a Monte Carlo simulation and rejection sampling to generate a set of instances of the preference model compatible with DM supplied indirect preference information.

%% file: applications.tex

\section{MOEA/D for Real-World Applications}
\label{sec:applications}

In addition to the prosperity of algorithmic developments, MOEA/D and its variants have shown competitiveness in many real-world applications. In this section, we overview some selected application cases of MOEA/D on machine learning, routing, engineering optimization, complex networks, and portfolio optimization.

\subsection{Applications on Machine Learning}
\label{sec:app_ml}

Machine learning tasks are essentially optimization problems that aim to identify optimal model parameters best fit the underlying data. Here we overview some selected applications of MOEA/D for different machine learning tasks.

\begin{itemize}
    \item\underline{\textit{Feature selection}}: As a fundamental step in machine learning and data mining tasks, feature selection aim to select a set of relevant features to facilitate predictive modeling and partially overcome the curse of dimensionality. It is a challenging problem due to the intractable interactions among features and multiple conflicting objectives. There have been some MOEA/D variants that aim to improve the model performance as well as minimizing the number of features~\cite{DemirNXZ20,NguyenXAIZ20,HanCLH21}.

    \item\underline{\textit{Neural architecture search (NAS)}}: This is an emerging area that aims to democratize deep learning at the fingertip of domain experts. Traditional gradient descent and reinforcement learning approaches mainly consider the validation accuracy as the only objective function in NAS. By considering objectives other than performance, such as efficiency and model complexity, some MOEA/D variants have been proposed for NAS where multiple trade-off architectures can be identified at a time~\cite{CaiGRMD15,LiuGMWL18,WuZLCC18,JiangHLWLH20}.

    \item\underline{\textit{Adversarial example generation}}: As opposed to NAS, adversarial example generation is another emerging area that aims to fool a trained neural network by perturbations. To achieve a balance between imperceptibility and attack capability, adversarial example generation is a typical MOP that aims to maximize the misclassification rate while minimize the distortion. Given the intractability of the deep neural networks under test, evolutionary computation has been recently recognized as a promising alternative for black-box attacks. There have been some initial attempts that apply MOEA/D to generate perturbations patterns on images~\cite{SuzukiTO19,DengZW19}.

    \item\underline{\textit{Association rule mining}}: It aims to find dependency relationship between items from large datasets~\cite{TelikaniGS20}. The process of extracting association rules can be formulated as a MOP that jointly optimizes various measures such as accuracy, comprehensibility, surprisingness and novelty. There have been some MOEA/D variants for solving multi-objective rule mining problem~\cite{ChanCF10} and identifying a reduced set of positive and negative quantitative association rules with a low computational cost~\cite{RodriguezRAH14}.
\end{itemize}

\subsection{Applications on Routing Problems}
\label{sec:app_routing}

Routing problems are ubiquitous in various areas including, but not limited to, transportation, logistics and communications. Typical objectives include routing distances and associated costs along with various constraints such as capacity and time window. Here we review some MOEA/D variants for solving different types of routing problems.
\begin{itemize}
    \item\underline{\textit{Travelling salesman problem (TSP)}}: Given a weighted and undirected complete graph, TSP is one of the most famous routing problems that aims to find the shortest Hamiltonian cycle covering all nodes~\cite{ApplegateBCC06}. Several MOEA/D variants have been developed mainly with dedicated reproduction operators (e.g., SA~\cite{LiS08,LiL11}, EDAs~\cite{ShimTC12} and ACO~\cite{KeZB14}) and local search strategies (e.g., guided local search~\cite{AlhindiAAATA19} and collaborative local serach~\cite{CaiXZMHWH21}). In addition to reproduction operators, \cite{CaiLFZ15} proposed a hybrid selection mechanism alternating between non-dominated sorting and decomposition for combinatorial optimization problems including TSPs.

    \item\underline{\textit{Vehicle routing problem (VRP)}}: As a generalization of TSP, vehicle routing problem (VRP) is a challenging combinatorial optimization and integer programming problem that aims to find optimal routes of multiple vehicles covering all nodes in a graph~\cite{JozefowiezST08}. Note that since all vehicles start from the same depot with a maximum capacity, VRP is more challenging than TSP given more complex encoding and stricter constraints of capacities. For example, new reproduction operators (e.g., chemical reaction optimization~\cite{LiWHLJ18}) and local search strategies (e.g., multi-mode mutation~\cite{GeeAJT16} and variable neighborhood descent~\cite{LanLNZG20}) have been proposed to generate more diversified solutions that meet capacity and time window constraints. In~\cite{WangWZ19}, a two-stage strategy is proposed to improve the ability of MOEA/D for finding better solutions in VRPs with multiple depots. Besides, tailored selection operators~\cite{QiHLHL15,Garcia-NajeraBG15} have been proposed to achieve a balance between convergence and diversity.

    \item\underline{\textit{Pickup and delivery problem (PDP)}}: Comparing to TSP and VRP, pickup and delivery problem (PDP) considers a more complex scenario consists of multiple pairs of pickup and delivery locations. In particular, each pickup location should be visited before the corresponding delivery location in the same route within a given time window~\cite{GrandinettiGPP14}. One of the major challenges of PDP is its highly discrete search space, MOEA/D variants with tailored local search strategies~\cite{WangZWZCZ16,WangSZG20} have been developed to mitigate this problem.

    \item\underline{\textit{Location routing problem (LRP)}}: As a integration of facility location and VRP, LRP is a challenging NP-hard problem that aims to determine the locations of depots and their corresponding vehicle routes to serve the customers under a series of constraints~\cite{SchneiderD17}. Recently, \cite{WangYZGSZ21} proposed a hybrid local search strategy including general local search, objective-specific local search and large variable search under the framework of MOEA/D.
\end{itemize}

\subsection{Applications to Engineering Optimization}
\label{sec:app_design}

Engineering optimization problems are full of multiple conflicting objectives given the multi-facet requirements to accommodate a diverse range of stakeholders. In the following paragraphs, we briefly overview some selected applications of MOEA/D according to the application domains.
\begin{itemize}
    \item\underline{\textit{Antenna optimization}}: It aims at creating advanced and complex electromagnetic devices that must be competitive in terms of performance, serviceability, and cost effectiveness under certain resource and communication constraints. MOEA/D variants have been applied in antenna design problems such as compact log-periodic dipole array antenna design~\cite{LiCCD19}, dual-polarized compact MIMO antennas~\cite{LuWWW17} and Yagi-Uda antennas~\cite{CarvalhoSGLM12}.

    \item\underline{\textit{Power systems}}: The planning, operation, and control problems in power systems are challenging optimization problems given their large-scale, complex and geographically widely distributed characteristics. MOEA/D and its variants has been applied to the planning problem of electric vehicle charging stations and power distribution system~\cite{WangDLMZ18}, electric motor design~\cite{SilvaLRL17}, optimal operation of hybrid energy systems~\cite{ZhangYYLY21}, power system voltage stability design~\cite{XuDMYZW14} and optimal power flow optimization~\cite{XiaoSMZ20}.

    \item\underline{\textit{Processs control}}: A MOEA/D variant equipped with tailored local search strategies was proposed to solve color-coating scheduling~\cite{DongWT20}, a mixed integer nonlinear programming problems.

    \item\underline{\textit{Search-based software engineering (SBSE)}}: Many activities in software engineering can be formulated as optimization problems. In particular, SBSE is an emerging area that mainly converts a software engineering problem into a computational search problem that can be tackled with a meta-heuristic. There have been several attempts of applying MOEA/D for solving software project scheduling problems~\cite{XiaoGH15}, self-adaptive software configurations~\cite{ChenLBY18}, software next release prediction~\cite{CaiCFGW17}, and software product line testing~\cite{FerreiraLSKVP17}.
\end{itemize}

\subsection{Applications to Complex Networks}
\label{sec:app_cn}

Fundamental features and intricate structures of many real-world complex systems can be modeled as complex networks. 
\begin{itemize}
    \item\underline{\textit{Network clustering and community detection}}: Network clustering is essential for understanding the community structure (also known as community detection) and functionality of a complex network. Different from clustering in the Euclidean space, network clustering is more challenging given its discrete characteristics, large-scale network size, and nontrivial measurements of distances between nodes. There have been some MOEA/D variants developed to tackle the discrete structure~\cite{GongCCM14,ZhangSJ20}, curse-of-dimensionality~\cite{ZhangZPZZJ20}, and distance measures~\cite{ChengCSNZ18}.

    \item\underline{\textit{Critical node detection}}: Similar to the sensitivity analysis but in a discrete space, this task aims to find the critical nodes in a complex network. One of the major challenges of this task is its plateau landscape as most node reductions do not split the network. To address this problem, \cite{LiPXC19} proposed a bi-objective formulation with dedicated mating and environmental selection mechanisms.
\end{itemize}

\subsection{Applications to Portfolio Optimization}
\label{sec:app_fintech}


Portfolio optimization is a classic problem in economics that usually aim to maximize profits while minimizing risks. It consists of two stages: optimizing weights of asset classes to hold, and optimizing weights of assets within the same asset class. One of the major challenges of this problem is the badly different scales with regard to different objectives. There have been some MOEA/D variants with a novel normalization method~\cite{ZhangLMT10}, weight adaptation method~\cite{ZhangZWZYS18} and multi-dimensional mapping coding scheme~\cite{ChenZ19} to address this problem.

%% file: emerging.tex

\section{Emerging Topics and Future Research Directions}
\label{sec:others}

Many real-world applications bring new challenges which lead to desirable research directions in the coming future. 

\begin{itemize}
    \item\underline{\textit{Large-scale MOPs (LSMOPs)}}: It is not uncommon that real-world optimization problems involve a large number of decision variables, also known as LSMOPs. Cooperative coevolution is one of the most popular and effective approaches for large-scale global optimization. Its basic idea is to decompose the original large-scale problem into several small-scale subproblems based on a decision variable analysis that captures the variable interaction. Then cooperative coevolution is applied to solve these subproblems in a collaborative manner. In the context of MOEA/D, the decision variable analysis (DVA) have also been the most popular technique for LSMOPs. As an early attempt, \cite{MaLQWLJYG16} proposed a DVA technique that separates decision variables based on their control property, i.e., being convergence- or/and spread- related. Thereafter, the proposed DVA technique separately interprets the interaction structures of both convergence- and spread-related variables separately based on random permutations.  This DVA technique is further extended as a graph representation in~\cite{CaoZGLM20}, being solved in a distributed manner~\cite{CaoZLL17} and an adaptive localized DVA~\cite{MaHYWW21}. Another interesting attempt is~\cite{QiBMML16} that uses a self-adaptive operator selection mechanism to enable MOEA/D to solve LSMOPs instead of analyzing variable interactions. Nevertheless, comparing the scale in the real-world scenarios, e.g., GPT-3 language model created by OpenAI contains 175 billion parameters~\cite{BrownMRSKDNSSAA20}, existing algorithmic developments of EMO for LSMOPs are still far from applicable, usually less than $10,000$ decision variables. To improve the scalability of EMO for LSMOPs to meet the industrial-level requirements, it is imperative to develop effective and efficient decomposition methods, such as novel problem transformation and dimension reduction techniques to narrow down the high-dimensional search space, so that EMO algorithms can approximate the optimal solutions with a manageable amount of function evaluations. Given the large-scale characteristics, it is also important to develop parallel computing platform by leveraging modern high-performance computing devices such as GPUs to improve the efficiency.

    \item\underline{\textit{Multimodal MOPs (MMOPs)}}: In some real-world optimization scenarios, such as space mission design problems~\cite{SchutzeVC11}, diesel engine design problems~\cite{HiroyasuNM05} and rocket engine design problems~\cite{KudoYF11}, there exist some dissimilar solutions in the decision space but lead to similar or equivalent quality in the objective space, as known as MMOPs. Solving an MMOP aims to locate (almost) equivalent Pareto optimal solutions as many as possible. To maintain the diversity of the population in both the objective and decision spaces, \cite{TanabeI18,TanabeI20a} proposed a general framework that uses the environmental selection mechanism of the baseline EMO algorithm to maintain diversity in the objective space while the diversity in the decision space is maintained by a simple niching criterion. Recently, a graph Laplacian based optimization assisted by decomposition in both decision and objective spaces is proposed in~\cite{PalB21} for solving MMOPs. More effective distance metrics and diversity preservation strategies are expected to better identify multiple equivalent Pareto-optimal solutions simultaneously for MMOPs.

    \item\underline{\textit{Variable-length MOPs}}: In conventional optimality theory, the length of decision variables is always assumed to be fixed. However, it is not uncommon that the optimal length of a decision variable is unknown \textit{a priori}, let alone the optimum, in many real-world applications, such as composite laminate stacking problem~\cite{SoremekunGHW01}, sensor coverage problem~\cite{TingLCW09} and wind farm layout problem~\cite{HerbertPRLC14}. This can lead to a variable-length Pareto structure where the number of variables in two Pareto-optimal solutions might be different. One of its key challenges is the varying length of variables that disrupts the genetic reproduction operators in EA. Furthermore, due to the existence of variable-length Pareto structure, approximating the PF can be biased as the local region with a larger variable size is more difficult than those with smaller variable sizes~\cite{LiD17}. To address these issues, MOEA/D-VLP~\cite{LiDZ19} proposed a two-level decomposition strategy where the global level is the same as the conventional MOEA/D while the local level decomposes an MOP into multiple MOPs with different ranges of dimensionality to approximate various local PF segments. In addition, an adaptive strategy is proposed to allocate computational efforts towards different parts of the PF to mitigate the bias towards any relatively easier PF regions. As a next step, it is desirable to develop bespoke subproblem formulations for different regions of the PF and a balanced way to strategically allocate computational budget.

    \item\underline{\textit{Sparse optimization}}: Sparse optimization problems widely appear in many real-world applications such as data mining, variable selection, visual coding, signal and image processing. Its ultimate goal is to find a sparse representation of a system, which usually aims to minimize a data-fitting term and a sparsity term simultaneously. As a dominating strategy in sparse optimization, the classic iterative thresholding often aggregates these two terms into a single function, where a relaxed parameter is used to balance the error and the sparsity. In particular, the choice of this relaxed parameter is sensitive to the performance of the iterative thresholding methods. There have been some initial attempts~\cite{LiSXZ12,LiSWZ18,LiZDX18} that transform the sparse optimization problem as a MOP and they apply tailored MOEA/D to solve it. Due to the unknown sparsity level of the underlying problem, it is challenging to maintain the diversity of the population. In addition, it is imperative to design bespoke performance indicator to measure the convergence, diversity, and sparsity of the population simultaneously.

    \item\underline{\textit{Parallel computing}}: The decentralized nature of decomposition makes it very appealing to use high-performance computing in order to distribute the underlying computations on the increasingly available compute facilities. While being a challenging issue due to the heterogeneous and complex nature of modern large-scale computing platforms, incorporating and enabling parallelism is to be tightly coupled with the other decomposition components and not in a separate manner. There have been very limited studies on the distributed implementation of MOEA/D based on the classic master-slave model~\cite{LiaoIPS20} and a GPU-based CUDA technology for real-time MOPs~\cite{YuLQ18}. As a next step, it is desirable to develop a mapping function that aligns the decomposed subproblems with the parallelism exposed by the computing platform. Moreover, it is necessary to design adaptive distributed algorithms in line with the characteristics of the subproblems.

    \item\underline{\textit{Theoretical studies}}: Although MOEA/D and its variants have been studied for solving various challenging MOPs, theoretical understanding of its behavior is far left behind. The current study mainly focus on understanding the relationship among some classic subproblem formulations as well as the influence of their corresponding hyper-parameters. For example, \cite{MaZTYZ18} theoretically investigated the geometric properties of the TCH subproblem formulation and found some asymptotically equivalence between the TCH function and the Hypervolume indicator. Singh and Deb~\cite{SinghD20} investigated the equivalence between the PBI and the AASF subproblem formulations under different hyper-parameter settings. An open research direction is to mathematically prove the convergence property of the population-based meta-heuristics under the framework of MOEA/D. To the best of author's knowledge, no work in this direction has been reported yet. In particular, the matching-based selection mechanism proposed in~\cite{LiZKLW14} is a good starting point to analyze the equilibrium of the evolutionary process.


\end{itemize}

%% file: conclusion.tex

\vspace{-1em}

\section{Conclusions}
\label{sec:conclusions}

MOEA/D opens up the pathway that bridges the gap between population-based EMO and the conventional MCDM. This article has presented a comprehensive survey of the up-to-date developments of MOEA/D. To be self-contained, we started with a gentle tutorial about the basic working mechanism of MOEA/D. Thereafter, our survey is conducted according to the four core design components, i.e., weight vector settings, subproblem formulations, selection mechanisms and reproduction operators, along with some selected advanced topics including constraint handling, expensive optimization and preference incorporation\cite{LiZZL09,LiZLZL09,CaoWKL11,LiKWCR12,LiKCLZS12,LiKWTM13,LiK14,CaoKWL14,WuKZLWL15,LiKD15,LiDZ15,LiDZZ17,WuKJLZ17,LiDY18,LiCSY19,Li19,GaoNL19,LiXT19,ZouJYZZL19,LiLDMY20,WuLKZ20,LiX0WT20}. At the end, we outlined some emerging directions for future developments of MOEA/D that have not yet been broadly studied so far. We believe that all these are expected to produce significant advances in the design and analysis of MOEA/D as well as its further uptake in real-world applications~\cite{ChenLBY18,ChenLY18,BillingsleyLMMG19,LiuLC20,LiXCT20}.

\textbf{MOEA/D Resources}: There are a series of technical activities themed on the decomposition multi-objective optimization. For example, there is a dedicated website\footnote{\url{https://sites.google.com/view/moead/}} gathering various information related to MOEA/D, including relevant papers, source codes and active researchers. Sponsored by the IEEE Computational Intelligence Society, there has been a Task Force themed on decomposition-based techniques in EC\footnote{\url{https://cola-laboratory.github.io/docs/misc/dtec/}} since 2017. This is an international consortium that brings together global researchers to promote an active state of this area. Since 2018, there have been tutorials, workshops and special sessions associated with major conferences in EC, including PPSN, GECCO, CEC and SSCI.